\documentclass[preprint,1p,times,nonatbib]{elsarticle}

\newgeometry{
  top=4.3cm,
  right=4.8cm,
  bottom=4.3cm,
  left=4.8cm
}

\makeatletter
\let\c@author\relax
\makeatother

\usepackage[backend=biber,style=ieee,url=false,natbib=true]{biblatex}
\usepackage{amsmath,amssymb,amsfonts}
\usepackage{algorithm}
\usepackage{algpseudocode}
\usepackage{graphicx}
\usepackage{textcomp}
\usepackage{booktabs}
\usepackage{pifont}
\usepackage{multirow}
\usepackage{boldline}
\PassOptionsToPackage{table}{xcolor}
\usepackage{xcolor}
\usepackage{colortbl}
\usepackage{color}
\usepackage{caption}
\usepackage{subcaption}
\usepackage[section]{placeins}
\usepackage{etoolbox}
\usepackage{hyperref}

\AtBeginBibliography{\raggedright}

\DeclareCaptionFont{eightpt}{\fontsize{8pt}{9.5pt}\selectfont}
\renewcommand{\footnotesize}{\fontsize{8pt}{9.5pt}\selectfont}

\AtBeginEnvironment{table}{\small}
\AtBeginEnvironment{tabular}{%
  \fontsize{7.5pt}{9pt}\selectfont
  \setlength{\tabcolsep}{2pt}%
}

\newcolumntype{C}[1]{>{\centering\arraybackslash}p{#1}}
\newcolumntype{L}[1]{>{\raggedright\arraybackslash}p{#1}}
\newcolumntype{R}[1]{>{\raggedleft\arraybackslash}p{#1}}

\makeatletter
\newcommand{\groupedRowColors}[5][0]{
    \global\rownum=\z@
    \global\@rowcolorstrue
    \@ifxempty{#4}%
        {\def\@oddrowcolor{\@norowcolor}}%
        {\def\@oddrowcolor{\gdef\CT@row@color{\CT@color{#4}}}}%
    \@ifxempty{#5}%
        {\def\@evenrowcolor{\@norowcolor}}%
        {\def\@evenrowcolor{\gdef\CT@row@color{\CT@color{#5}}}}%
    \def\@rowcolors{%
        \if@rowcolors
            \noalign{%
                \relax
                \ifnum\rownum<#3
                    \@norowcolor
                \else \ifodd \numexpr (\rownum-#1)/#2\relax
                    \@oddrowcolor
                \else
                    \@evenrowcolor
                \fi \fi
            }%
        \fi
    }%
    \CT@everycr{\@rowc@lors\the\everycr}%
    \ignorespaces
}
\makeatother

\begin{document}

\begin{frontmatter}

\title{Generalized Design Choices for Deepfake Detectors\tnoteref{repo}}
\tnotetext[repo]{Project repository: \url{https://github.com/MI-BioLab/AI-GenBench}.}

\author[inst1]{Lorenzo Pellegrini\corref{cor1}}
\ead{l.pellegrini@unibo.it}
\author[inst1]{Serafino Pandolfini}
\author[inst1]{Davide Maltoni}
\author[inst1]{Matteo Ferrara}
\author[inst2]{Marco Prati}
\author[inst2]{Marco Ramilli}

\cortext[cor1]{Corresponding author}
\address[inst1]{Department of Computer Science and Engineering (DISI), University of Bologna, Italy}
\address[inst2]{IdentifAI, Italy}

\begin{abstract}
The effectiveness of deepfake detection methods often depends less on their core design and more on implementation details such as data preprocessing, augmentation strategies, and optimization techniques. These factors make it difficult to fairly compare detectors and to understand which factors truly contribute to their performance. To address this, we systematically investigate how different design choices influence the accuracy and generalization capabilities of deepfake detection models, focusing on aspects related to training, inference, and incremental updates. By isolating the impact of individual factors, we distinguish robust cross-backbone trends from architecture-dependent effects and derive practical recommendations for future deepfake detection systems. Our experiments identify a set of design choices that consistently improve deepfake detection and enable state-of-the-art performance on the AI-GenBench benchmark.
\end{abstract}


\end{frontmatter}

\section{Introduction}
The rapid advancement of generative models has led to an unprecedented ability to produce realistic synthetic images that are increasingly difficult to distinguish from human-generated (real) content. Diffusion-based architectures and large-scale text-to-image systems such as Stable Diffusion, Midjourney, and DALL-E have significantly lowered the barrier to generating high-quality content. In particular, this has enabled professionals as well as the general public to create new media content conditioned by a text prompt and other semantic inputs. These technologies have often been misused to disseminate disinformation, raising societal concerns and establishing the detection of AI-generated content as an important research area \cite{epstein2023art, barrett2024identifying, lin2024detecting}, which is essential for preserving trust, accountability, and authenticity in digital media.

\begin{figure}[t]
    \centering
    \includegraphics[width=0.9\linewidth,
    trim= 90 200 140 200,
    clip]{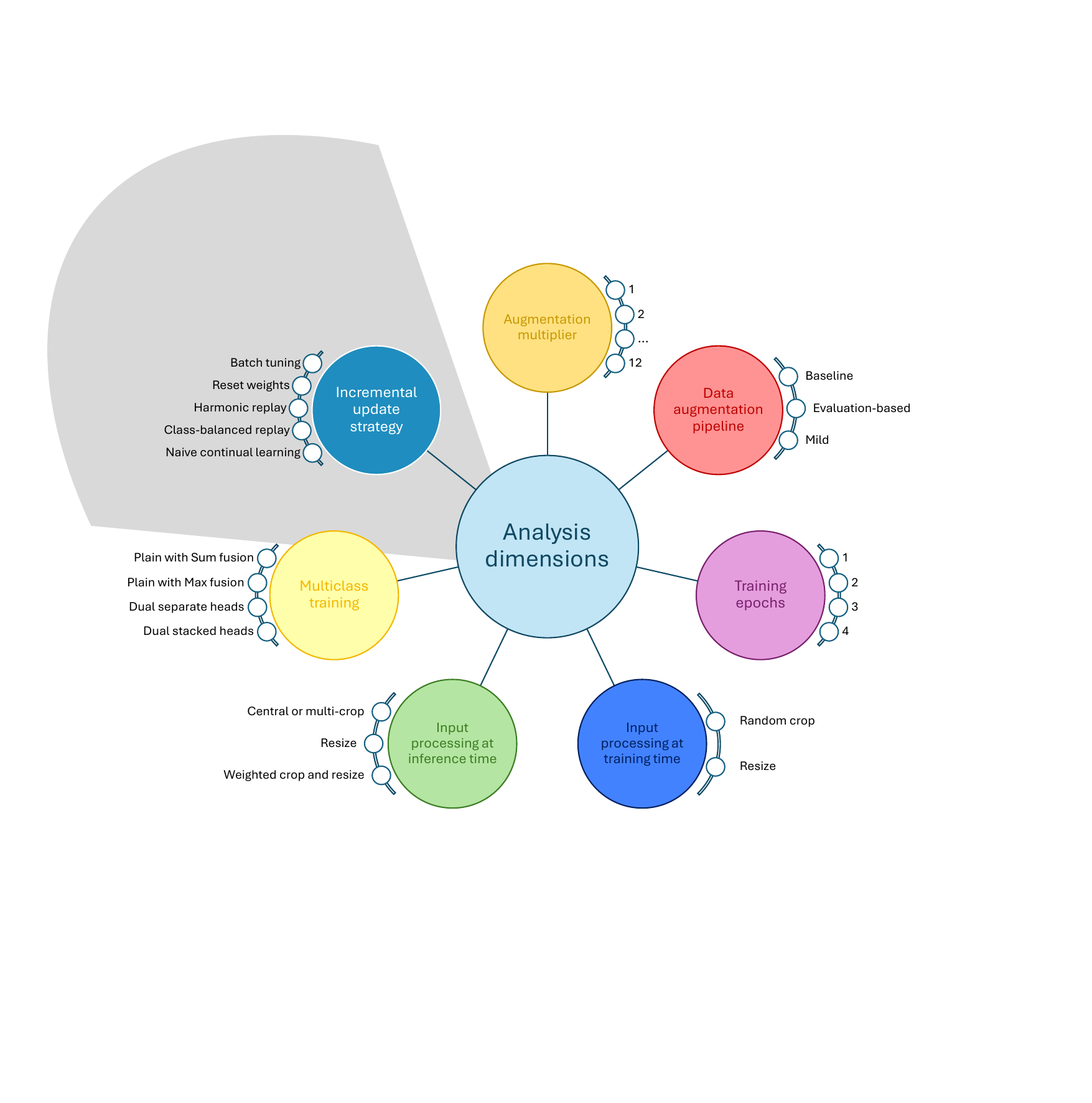}
    \caption{The different dimensions explored in this work, to optimize the training, inference and incremental update (grayed) of deepfake detectors.}
    \label{fig:teaser}
\end{figure}

Over the past few years, numerous detection approaches have been proposed, ranging from handcrafted forensic cues to deep neural networks specifically trained to distinguish real from synthetic content. Despite promising progress, detection performance often varies widely across studies, datasets, and model architectures. In many cases, the reported success of a particular method depends less on the core detection idea than on specific, and sometimes implicit, implementation details—such as the choice of data augmentations, preprocessing, or training strategy. This lack of systematic evaluation makes it difficult to identify which design factors truly contribute to generalization across different types of generators and model architectures. Moreover, existing works often focus on training detection models on content produced by a limited (or even a single) set of hand-picked generators and then testing such models on images from other generators.

In this work, we present a comprehensive empirical study aimed at isolating general design principles from the influence of specific model architectures or fake data generators. To this purpose, we adopt the recent AI-GenBench benchmark~\cite{pellegrini2025ai}, which temporally orders image generators to simulate the release of new generative models over time. With this setup, we systematically evaluate how various training and inference-time choices affect the generalization capability of detection models on both old (such as GAN-based) and recent generative techniques. Specifically, we analyze the impact of (i) the data augmentation pipeline, (ii) the training duration and augmentation multiplier, (iii) the preprocessing strategies such as cropping versus resizing, and (iv) the use of multiclass labels and related strategies. As a further dimension of analysis, we consider the incremental training of deepfake detectors. This is a relevant issue for the practical deployment of detectors that must be frequently updated to cope with novel and emerging generation techniques. Since retraining from scratch on a steadily increasing dataset can be very resource-consuming, we evaluate how detection models can be trained incrementally in a sample-efficient way while preserving detection capabilities on older generators, a scenario commonly referred to as \textit{Continual Lifelong Learning}. The study is therefore organized around practical questions: which design choices improve future-generator generalization, which effects depend on the backbone, and which gains remain attractive once computational cost is considered. An overview of the directions explored in this work is reported in Figure~\ref{fig:teaser}.

Our goal is not to introduce a new detection method, but rather to identify which recommendations remain stable across backbones and which should instead be treated as model-specific observations. To this end, we systematically evaluate multiple pre-trained vision backbones such as ResNet-50 CLIP, ViT-L CLIP, DINOv2, and EfficientNet-B0. EfficientNet-B0 provides a lightweight CNN reference point, while DINOv2 ViT-L covers the large self-supervised regime; we do not extend the study to even larger backbones because this would substantially increase computational cost and reduce comparability with the ViT-L-scale models commonly adopted in recent detection studies~\citep{ojha2023univfd, yermakov2026deepfake, nguyen2024exploring}. The findings of this study offer actionable insights for the development of future detection systems, equipping both researchers and practitioners with a solid foundation for designing robust detectors of AI-generated images. To the best of our knowledge, this is the first model-agnostic study that systematically evaluates all those dimensions of deepfake detection, covering both training and evaluation mechanisms.

This work is organized as follows. Section~\ref{sec:related} introduces the background and reviews relevant related work. Section \ref{sec:experimental_setup} describes the experimental setup, detailing the adopted benchmark and the main dimensions of analysis. Section \ref{sec:results} presents the results of these experiments. We discuss Incremental Update Strategies in Section \ref{sec:ius}, as this topic is mostly orthogonal to the other dimensions. Section \ref{sec:best_of} reports the detection performance achieved by the ``best of'' configuration, which combines the most effective approaches identified in our analysis. Finally, Section \ref{sec:conclusions} summarizes our findings and outlines future research directions.

\section{Related work}
\label{sec:related}
Deepfake detection aims to distinguish AI-generated content from authentic media including images, videos, and audio. In this section, we review relevant literature in two key areas of deepfake image detection: (i) detection methods, which focus on the backbones and algorithmic strategies used to identify manipulated content, and (ii) benchmarks, which provide datasets and evaluation protocols essential for developing, validating, and comparing detection techniques. A final subsection is devoted to a brief review of relevant continual learning techniques.

\subsection{Detection Methods}
Early methods for differentiating synthetic from authentic images predominantly relied on Convolutional Neural Networks (CNNs) trained on large-scale datasets \cite{lin2024detecting}. While these approaches achieve high accuracy under conditions closely aligned with the training distribution, their performance degrades significantly in real-world scenarios. In particular, they tend to be vulnerable to common image degradations such as compression, resizing, blurring, and cropping that frequently occur when images are shared via social media or instant messaging platforms. Recent online-social-network-oriented approaches explicitly address this issue by learning from compressed and unpaired data or by reducing the influence of compression block artifacts \cite{tao2025oddn,li2025plada}. In such scenarios, detection systems often struggle to generalize to images generated by previously unseen models \cite{Tariang2024synthetic}. To mitigate these limitations, incorporating carefully designed data augmentation strategies during training has proven effective. These augmentations not only enhance robustness against image-level perturbations but also improve cross-generator generalization \cite{wang2020cnn}. Consequently, the design of the training augmentation pipeline is one of the main elements investigated in our study. While ImageNet-pretrained CNN backbones dominated early research in this field, recent work has explored large models based on different structures, such as Vision Transformers, and different pre-training strategies, such as vision-language models like CLIP. These demonstrate strong performance even when trained on data from a single generator, thanks to their rich feature representations and superior transferability \cite{ojha2023univfd}. Furthermore, recent studies suggest that foundational vision backbones such as DINOv2, trained using self-supervised learning, may be particularly effective for deepfake detection \cite{pellegrini2025ai}. Other recent methods pursue the same generalization goal through semantic-agnostic artifact learning \cite{tao2025sagnet} or multimodal/foundation-model supervision \cite{tao2025lednet}, providing complementary directions to the implementation-focused analysis conducted here.

Among the plethora of methods proposed in the literature, some of them focus on the modification of network architectures to better capture low- and high-level forensic traces \cite{koutlis2024rine, sarkar2024shadows}, while others improve training strategies \cite{baraldi2024code,boychev2025imaginet} or simulate generator-specific artifacts \cite{Rajan2024effectiveness,guillaro2024bfree}. Recent research has also explored formulating the detection task beyond traditional binary or multiclass classification. For instance, LASTED \cite{wu2025generalizable} employs a language-guided contrastive learning objective to align images with descriptive text prompts, thereby learning representations that generalize more effectively to unseen generators. Another strategy for improving generalization involves few-shot or incremental learning methods \cite{Laiti2024conditioned,li2023continual,tian2024dynamic}. While promising, these approaches require access to images from generators, which may not always be available in the most challenging scenarios. An alternative line of research considers periodically retraining detectors while preserving the temporal order of generator releases \cite{epstein2023online}.
This approach leverages forensic traces from known generators, which are often similar to those in newer models. Indeed, it is reasonable to believe that artificial fingerprints \cite{corvi2023intriguing} from one generator can enable classifiers to generalize across entire families of models, not just individual ones \cite{wang2020cnn}.

\subsection{Benchmarks}
Early benchmarks for synthetic image detection primarily focused on GAN-based generators and were often limited to specific domains, such as facial imagery—e.g., ForgeryNet~\cite{he2021forgerynet}, DiffusionFace~\cite{chen2024diffusionface}, and DIFF~\cite{cheng2024diffusion}—or artistic content~\cite{wang2023benchmarking}. To overcome these domain-specific limitations, recent benchmarks emphasize generalization by introducing large-scale, diverse datasets~\cite{rahman2023artifact, zhu2024genimage, hong2024wildfake, boychev2025imaginet}, or by providing open-source frameworks that facilitate the integration and evaluation of new generative models~\cite{schinas2024sidbench}. Additional efforts include comprehensive evaluations of existing datasets~\cite{park2024performance}, as well as human perceptual studies on synthetic content detection~\cite{lu2024seeing}. Moreover, several influential datasets, although not explicitly designed as benchmarks, are widely adopted by the research community for training and evaluation purposes. These include both GAN-based~\cite{wang2020cnn} and diffusion-based image collections~\cite{ojha2023univfd, corvi2023detection, bammey2023synthbuster, cazenavette2024fakeinversion}.

A common evaluation protocol across many benchmarks is to assess the generalization capability of detection models by testing them on generators unseen during training. However, this setup often overlooks the temporal evolution of generative techniques. In practice, new architectures are released continuously, making it essential to evaluate generalization under temporally realistic conditions: training on older generators and testing on newer ones according to their historical release timeline. This perspective was first introduced in~\cite{epstein2023online}, which demonstrated a significant drop in detection performance when models encountered major shifts in generative architectures. Building on this insight, AI-GenBench~\cite{pellegrini2025ai} introduced a temporal evaluation benchmark comprising 36 mainstream generative models released between 2017 and 2024. In our experiments, we adopt AI-GenBench as it provides a more realistic and forward-looking framework for assessing generalization over time. This benchmark is based on a protocol that defines the temporal order in which the generators are encountered. In addition, to allow for a fair comparison of different approaches, it also defines rules regarding the augmentation intensity to be used during training, the evaluation pipeline (and especially the augmentations used to introduce social media-like distortions), and the metrics used to evaluate the ability of each model to both generalize to unseen generators and retain detection capabilities on older ones. More information on the experimental protocol will be given in Section~\ref{sec:experimental_setup}.

\subsection{Continual learning}
Continual learning addresses the challenge of updating models with new data over time without losing performance on previously learned tasks, a problem commonly known as catastrophic forgetting~\cite{mccloskey1989catastrophic}. Existing approaches can be broadly categorized into three groups: (i) regularization-based methods, which constrain weight updates to preserve prior knowledge; (ii) architectural methods, which expand the model to allocate capacity for new tasks; and (iii) replay-based (or rehearsal) strategies, which retain and interleave a subset of past samples during training. Among these, replay methods are particularly popular due to their simplicity and effectiveness. They typically rely on memory buffers with various sampling and replacement policies~\cite{rebuffi2017icarl,chaudhry2019tiny}. In this work, we adopt replay-based strategies as a practical mechanism to mitigate forgetting in deepfake detection models, enabling adaptation to new generators without retraining on the entire data of past generators.

\section{Experimental setup}
\label{sec:experimental_setup}
The evaluation is conducted using the AI-GenBench temporal framework. In this benchmark, the 36 image generators are ordered by release date and split into temporal windows, each containing four generators. The detection model is trained progressively: at each step $k$, the model is trained on all generators within the sliding windows $w_j, j \leq k$. This setup simulates a realistic scenario where detectors are periodically retrained to keep up with novel generative models. After each training step $k$, the model is evaluated to measure its ability to detect images from both past and future generators. The benchmark defines a set of three scenarios on which the relevant metrics are measured:

\begin{itemize}
\item \emph{Next Period} - the detection performance is measured on the generators of the next sliding window ($w_{k+1}$).
\item \emph{Past Period} - performance is measured on the generators belonging to windows $w_j, j \leq k$.
\item \emph{Whole Period} - performance is measured on the generators belonging to both the past and next time windows ($w_j, j \leq k + 1$).
\end{itemize}

The benchmark proposes the \textit{Area Under Receiver Operating Characteristic Curve (AUROC)} as the main metric, which is averaged across all steps to obtain a single compact value. The performance measured on the \emph{Next Period} is particularly important as it measures the detector's ability to generalize to unseen generators, which will become available in the near future. For this reason, the authors of AI-GenBench consider the \emph{average AUROC on the Next Period} as the main metric.

Unless otherwise stated, every experiment is repeated with three independent random seeds, and all results are reported as the mean and standard deviation over the three runs, which captures the variability due to random initialization and stochastic training. Throughout the paper, figures and tables therefore report the three-run mean $\pm$ standard deviation. To additionally quantify the uncertainty due to finite test-set sampling, we computed nonparametric bootstrap confidence intervals over the test set using the stored per-item prediction scores: for each model, test instances were sampled with replacement and the evaluation metric recomputed over $10000$ bootstrap replicates, and the 2.5th and 97.5th percentiles are reported as $95\%$ confidence intervals. These bootstrap intervals are computed on a single run and are reported, for completeness, in Appendix~\ref{app:bootstrap}.

To identify which strategies generalize across different families of detectors, we consider the following well-known pre-trained vision (and language-vision) models: i) \textit{ResNet-50 CLIP} by OpenAI \cite{radford2021learning}, ii) \textit{ViT-L/14 CLIP} from LAION models\footnote{\href{https://huggingface.co/laion/CLIP-ViT-L-14-CommonPool.XL-s13B-b90K}{laion/CLIP-ViT-L-14-CommonPool.XL-s13B-b90K}}, iii) \textit{ViT-L/14 DINOv2} \cite{oquab2024dinov}, and iv) \textit{EfficientNet-B0} \cite{tan2019efficientnet}.
The EfficientNet-B0 backbone extends the analysis to a lightweight CNN family, while the DINOv2 ViT-L model provides the large self-supervised reference considered in this study.
We focus on the following design dimensions:

\begin{itemize}
\item \emph{Data augmentation pipeline} - evaluating the impact of transformations such as color jitter, Gaussian noise, blurring, geometric transformations, and especially JPEG compression.
\item \emph{Augmentation multiplier} - given an augmentation pipeline, systematically varying the number of diverse images presented to the model.
\item \emph{Training duration} - determining the optimal number of training epochs, for a given augmentation pipeline and multiplier.
\item \emph{Input processing at training time} - comparing training strategies based on image crops versus resized full images.
\item \emph{Input processing at inference time} - evaluating whether binary predictions are best obtained by (i) fusing scores from multiple image crops, (ii) using a resized version of the full image, or (iii) computing a weighted score from both multiple crops and (resized) full images.
\item \emph{Multiclass training} - investigating whether training the detection model on a multiclass problem using generator labels improves binary detection performance.
\begin{itemize}
    \item \emph{Multiclass to binary} - strategies to fuse multiclass scores into a binary prediction.
    \item \emph{Multiclass and Binary training} - assessing the benefits of training the model using both the multiclass and binary losses with different multi-head approaches.
    \item \emph{MLP vs distance-based approach} - exploring whether replacing the MLP classification head with a distance-to-centroid scoring function improves evaluation-time robustness after multiclass training.
\end{itemize}
\end{itemize}

Following the AI-GenBench protocol, we adopt the AUROC on the \emph{Next Period} (averaged across all steps) as the primary evaluation metric because it captures the detector's ability to generalize to future, unseen generators.

\subsection{Data augmentation pipeline}

Data augmentation plays a central role in enhancing the robustness of deepfake detection models, as it helps detectors generalize across different synthetic image generators and remain resilient to realistic image corruptions~\cite{wang2020cnn, mandelli2022detecting, gragnaniello2021are}. Robust feature-learning and transformation-invariant representations have also been studied in neighboring computer-vision and watermarking settings \cite{liu2024humanparsing,wang2025lightfield}, further supporting the broader motivation for explicitly training to be robust against strong transformations. To systematically assess its impact, we evaluate three distinct training-time augmentation pipelines, while at evaluation time all models are tested using the mandatory AI-GenBench preprocessing pipeline.

\subsubsection*{Baseline pipeline} 
The baseline pipeline is identical to the default augmentation strategy used in the AI-GenBench paper and initially proposed by Corvi et al. in~\cite{corvi2023detection}. It applies relatively strong transformations in a probabilistic manner, including random resized cropping, color jitter, grayscale conversion, dropout, Gaussian noise, blurring, random rotations, and horizontal flipping. A single JPEG compression pass is also applied with quality uniformly sampled from the range $[30,100]$. This configuration aims to improve generalization by exposing the detector to a wide spectrum of perturbations.

\subsubsection*{Evaluation-based pipeline} 
The second training pipeline is derived from the AI-GenBench evaluation pipeline, which was originally designed to simulate realistic degradations caused by upload, download, and re-encoding processes on social media or messaging platforms. This pipeline applies up to three successive JPEG compression passes with variable quality levels, combined with a softer set of augmentations compared to the baseline. For training purposes, we extend this pipeline by adding random horizontal flipping and random rotation. This design allows us to test whether training with more realistic and less aggressive augmentations can improve generalization while preserving robustness to real-world corruptions.

\subsubsection*{Mild pipeline} 
The third training pipeline is also derived from the AI-GenBench evaluation pipeline but, similar to the baseline pipeline, applies only a single final JPEG compression pass. The purpose of this intermediate configuration is to isolate the effect of repeated JPEG compression during training and determine whether multiple compression passes provide additional benefits compared to a simpler single-pass strategy. At evaluation time, all models are tested exclusively using the mandatory AI-GenBench evaluation pipeline, independently of the training pipeline used.

\subsection{Augmentation multiplier and training duration}
The augmentation multiplier ($am$) is a key hyperparameter introduced in the AI-GenBench framework to control the diversity of augmented images during training. Specifically, $am$ determines the number of unique augmented variants generated for each training image through deterministic augmentations. For instance, with $am=4$ (the default setting in AI-GenBench), the effective size of the training dataset becomes $4 \times |D|$, where $|D|$ denotes the number of original training images.

In the original AI-GenBench setup, training is performed for a single epoch with $am=4$, meaning the model sees exactly four distinct augmentations of each image. In our evaluation, we extend this analysis along two axes:

\begin{itemize}
    \item \emph{Varying augmentation multiplier} - we vary $am$ in the range $[1, 16]$ while keeping the number of epochs fixed at one. This isolates the effect of increasing augmentation diversity within a single pass over the dataset.
    \item \emph{Varying number of epochs} - we vary the number of epochs in the range $[1, 4]$ while fixing $am=4$. This isolates the effect of repeated passes over the augmented dataset while keeping augmentation diversity constant.
\end{itemize}

This setup enables us to disentangle the contribution of dataset expansion through augmentation from that of extended training duration, and to determine whether one or both factors are required to achieve optimal generalization performance.

\subsection{Input processing at training and inference time}
An important design choice for deepfake detection models concerns how input images are processed before being fed to the backbone. At training time, we consider two main strategies:

\begin{itemize}
    \item \emph{Random crop} - a sub-region of the image is randomly cropped to match the model’s input resolution. This strategy encourages the model to rely on fine-grained local artifacts and noise patterns that may reveal synthetic content.
    \item \emph{Resize} - the entire image is resized to the model’s input resolution, preserving global context. This allows the model to focus on semantic consistency and macroscopic distortions rather than local noise. However, severe downsizing may suppress subtle forensic cues.
\end{itemize}

At evaluation time, random crops are replaced by deterministic procedures:
\begin{itemize}
    \item \emph{Central crop or multi-crop} - either a single central crop or multiple crops followed by score fusion, approximating the training distribution of crop-based models.
    \item \emph{Resize} - the full image is resized, mirroring the training setup of resize-based models.
\end{itemize}

In the original AI-GenBench paper, crop-trained models were evaluated using the multi-crop strategy (with single-crop also tested but found inferior), while resize-trained models were evaluated on resized images. Their findings suggest that the resize strategy generally yields superior performance.

We extend this analysis by evaluating both crop- and resize-trained models under both inference protocols. Specifically, for each trained model we generate predictions from multiple crops and from the resized image, then fuse the scores with equal weight. Importantly, this \emph{Mixed} evaluation is applied separately to each model type: a crop-trained detector is never combined with a resize-trained detector. This setup allows us to test whether jointly leveraging local (crop-based) and global (resize-based) evidence at inference time improves robustness compared to relying on a single strategy. Finally, it should be noted that AI-GenBench contains images from multiple real and synthetic sources, and the native image sizes are therefore heterogeneous before the benchmark preprocessing is applied. From this perspective, the crop versus resize experiments should thus be interpreted as a controlled test of how detectors handle this resolution variability under the AI-GenBench protocol.

\subsection{Multiclass training}
Deepfake detection is typically formulated as a binary classification task: real versus synthetic. However, since training data often includes generator-specific labels, it is natural to ask whether reframing the problem as a multiclass task (real + $N$ generator classes) can improve binary detection performance. We investigate this question by evaluating both pure multiclass training and joint multiclass-binary approaches. We use generator identities only as training supervision. In fact, at inference time the generator identity is generally not available. Hierarchical supervision based on generator families (such as GAN-based, Diffusion-based, $\ldots$) could still be useful during training. We leave this enhancement option for future work.


\subsubsection*{Plain multiclass training}
In the first setting, we train the detector solely on the multiclass task, using the generator identity as the label. At evaluation time, the multiclass outputs must be converted into a binary prediction. We consider two fusion strategies:

\begin{itemize}
    \item \emph{Sum fusion} - the binary \textit{fake} score is computed by summing the softmax scores of all fake classes (with class $0$ representing the ``real'' class) encountered during training. 
    \item \emph{Max fusion} - the \textit{fake} score is defined as the maximum softmax score among all fake classes encountered during training.
\end{itemize}

\subsubsection*{Dual-head training}
In the second setting, we jointly train the model on both binary and multiclass objectives to assess whether generator-aware supervision can provide more discriminative features and whether enforcing both tasks jointly improves the final binary detection performance. To this end, we equip the backbone with two output heads and employ a combined loss retaining only the binary head at evaluation time. We experiment with two architectural variants:
\begin{itemize}
    \item \emph{Separate heads} - the backbone features two separate heads, one for binary classification and one for multiclass classification, trained simultaneously.
    \item \emph{Stacked heads} - the binary head is placed on top of the multiclass head; specifically, multiclass logits are passed through a ReLU and then projected onto a binary prediction.
\end{itemize}

For both variants, we explore different loss weightings: equal weighting ($0.5$ each) and an asymmetric configuration where the binary loss dominates ($0.75$ binary, $0.25$ multiclass), treating multiclass supervision as an auxiliary signal.

\subsubsection*{MLP vs distance-based approach}
In addition to standard multiclass-to-binary fusion, we explore a distance-based alternative to the usual MLP classification head. The procedure consists of five steps:

\begin{enumerate}
    \item \emph{Training} - the model is trained using the plain multiclass setup described above, without any dual-head architecture or binary loss.
    \item \emph{Centroid extraction} - for each class (generator), we extract $c$ centroids from the training set, with $c \in [1,3]$, to assess whether multiple centroids improve performance. Centroids are computed in the feature space immediately before the final classification layer. For $c > 1$, centroids are obtained by clustering training patterns using K-Means.
    \item \emph{Distance scoring} - for a test image, we compute its distance to each centroid of every class and convert this into an inverse-distance score, so that closer proximity corresponds to higher confidence.
    \item \emph{Class-level aggregation} - for each class, we retain the maximum inverse-distance score among its centroids.
    \item \emph{Binary prediction} - to produce a real/fake score, we apply a fusion strategy across the $N$ fake classes, analogous to the multiclass-to-binary approaches:
    \begin{itemize}
        \item \emph{Sum fusion} - sum the scores of all fake classes.
        \item \emph{Max fusion} - take the maximum score among all fake classes.
    \end{itemize}
\end{enumerate}

\subsection{Baseline}
All results are reported relative to a \emph{Baseline} configuration. This setup follows the procedure proposed in the initial AI-GenBench experiments and consists of using the \emph{Baseline pipeline} for data augmentation, training for a single epoch with an augmentation multiplier of $am=4$, resizing the entire image to the model's input size for both training and evaluation, and directly optimizing the binary classification objective (i.e., using only the binary loss).

\section{Results}
\label{sec:results}
The results are presented by design dimension. Each subsection first reports the empirical trend, then states whether the effect appears stable across backbones or should be interpreted as architecture-dependent.

\subsection{Data augmentation pipeline}

\begin{figure}[H]
\centering
\includegraphics[width=0.78\textwidth]{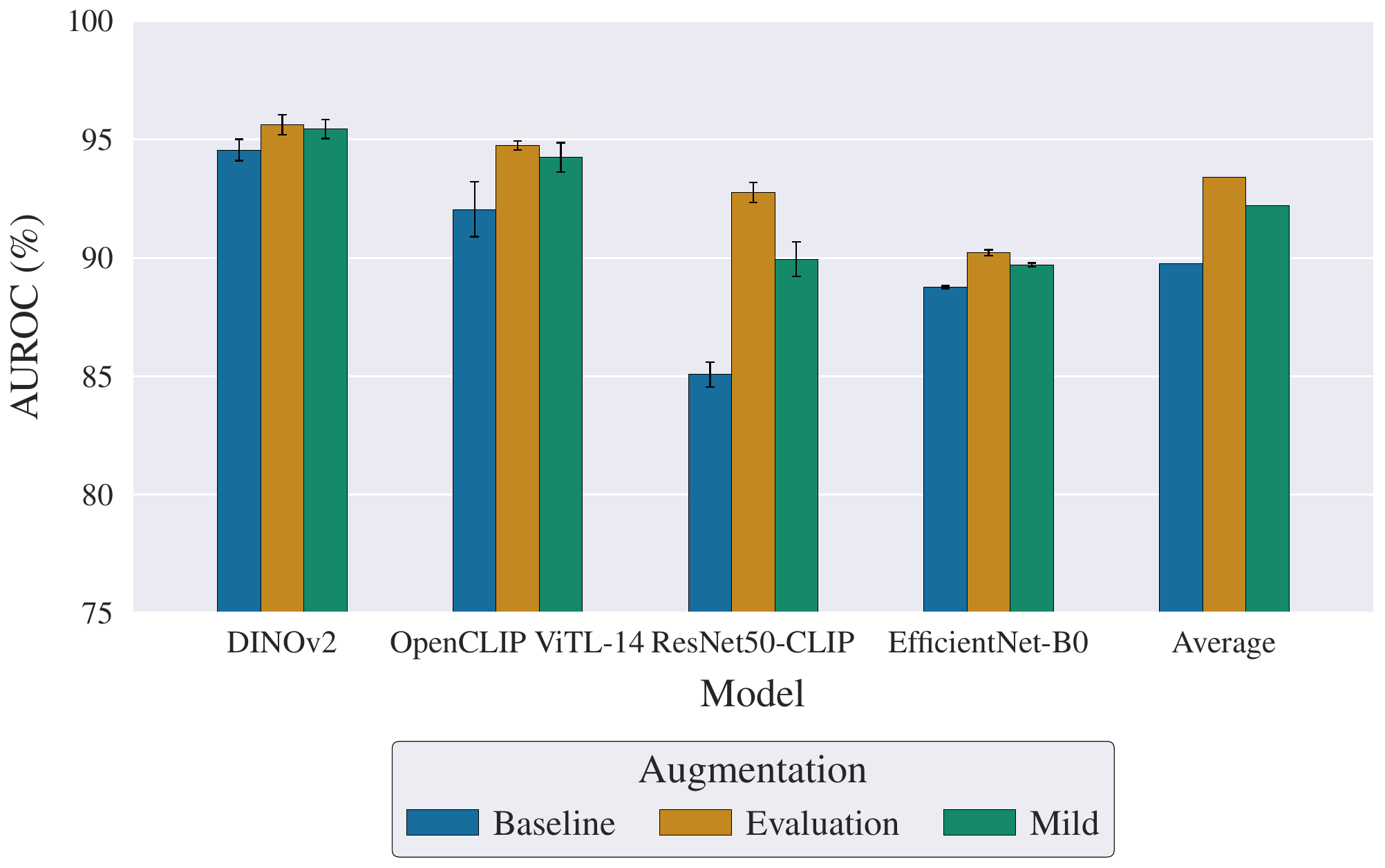}
\caption{Augmentation pipeline. Error bars denote the standard deviation across the three runs.}
\label{fig:augmentation_variation}
\end{figure}

Figure~\ref{fig:augmentation_variation} reports the performance of the three training-time augmentation pipelines across all considered backbones. While the \emph{baseline} pipeline, characterized by heavy perturbations and a single JPEG compression pass, achieves competitive results, we observe that the \emph{evaluation-based} pipeline (which applies up to three JPEG compression passes combined with milder augmentations) consistently leads to higher AUROC scores on the Next Period metric ($90.11\%$ vs. $93.34\%$ on average).

\begin{table}[H]
\centering
\small
\setlength{\tabcolsep}{3.5pt}
\begin{tabular}{lccccc}
\toprule
\textbf{Pipeline} & \textbf{DINOv2} & \textbf{ViT-L CLIP} & \textbf{ResNet-50 CLIP} & \textbf{EfficientNet-B0} & \textbf{Average} \\
\midrule
Baseline & 94.56 $\pm$ 0.45 & 92.05 $\pm$ 1.16 & 85.08 $\pm$ 0.52 & 88.76 $\pm$ 0.06 & 90.11 \\
Evaluation & \textbf{95.63 $\pm$ 0.42} & \textbf{94.76 $\pm$ 0.19} & \textbf{92.77 $\pm$ 0.43} & \textbf{90.22 $\pm$ 0.12} & \textbf{93.34} \\
Mild & 95.44 $\pm$ 0.40 & 94.24 $\pm$ 0.62 & 89.95 $\pm$ 0.72 & 89.71 $\pm$ 0.08 & 92.33 \\
\bottomrule
\end{tabular}
\caption{Effect of the training-time augmentation pipeline on the Next Period AUROC averaged over all temporal evaluation steps (\%). Values are mean $\pm$ standard deviation over the three runs; the Average column reports the mean across the different backbones. The best result per backbone (highest mean) is highlighted in bold. Bootstrap 95\% confidence intervals are reported in Appendix~\ref{app:bootstrap}.}
\label{tab:augmentation_variation}
\end{table}

The \emph{mild} pipeline, derived from the evaluation-based strategy but restricted to a single JPEG compression pass, achieves intermediate performance ($92.33\%$), suggesting that repeated compression during training plays an important role in preparing detectors for real-world degradations. Overall, these results indicate that:

\begin{itemize}
\item while data augmentation is critical for robust detection, excessively strong augmentations (as in the baseline pipeline) may be counterproductive;
\item augmentations that closely mimic realistic post-processing operations encountered in-the-wild provide more consistent improvements;
\item introducing repeated JPEG compression passes during training effectively improves the generalization capabilities, although for DINOv2 the advantage over the single-pass \emph{mild} pipeline ($95.63\%$ vs. $95.44\%$) is within the standard deviation across runs. Overall, these trends are observed across the evaluated detector architectures, underscoring the generality of this finding.
\end{itemize}

\subsection{Augmentation Multiplier and Training Duration}
\begin{figure}[H]
\centering
\includegraphics[width=0.78\textwidth]{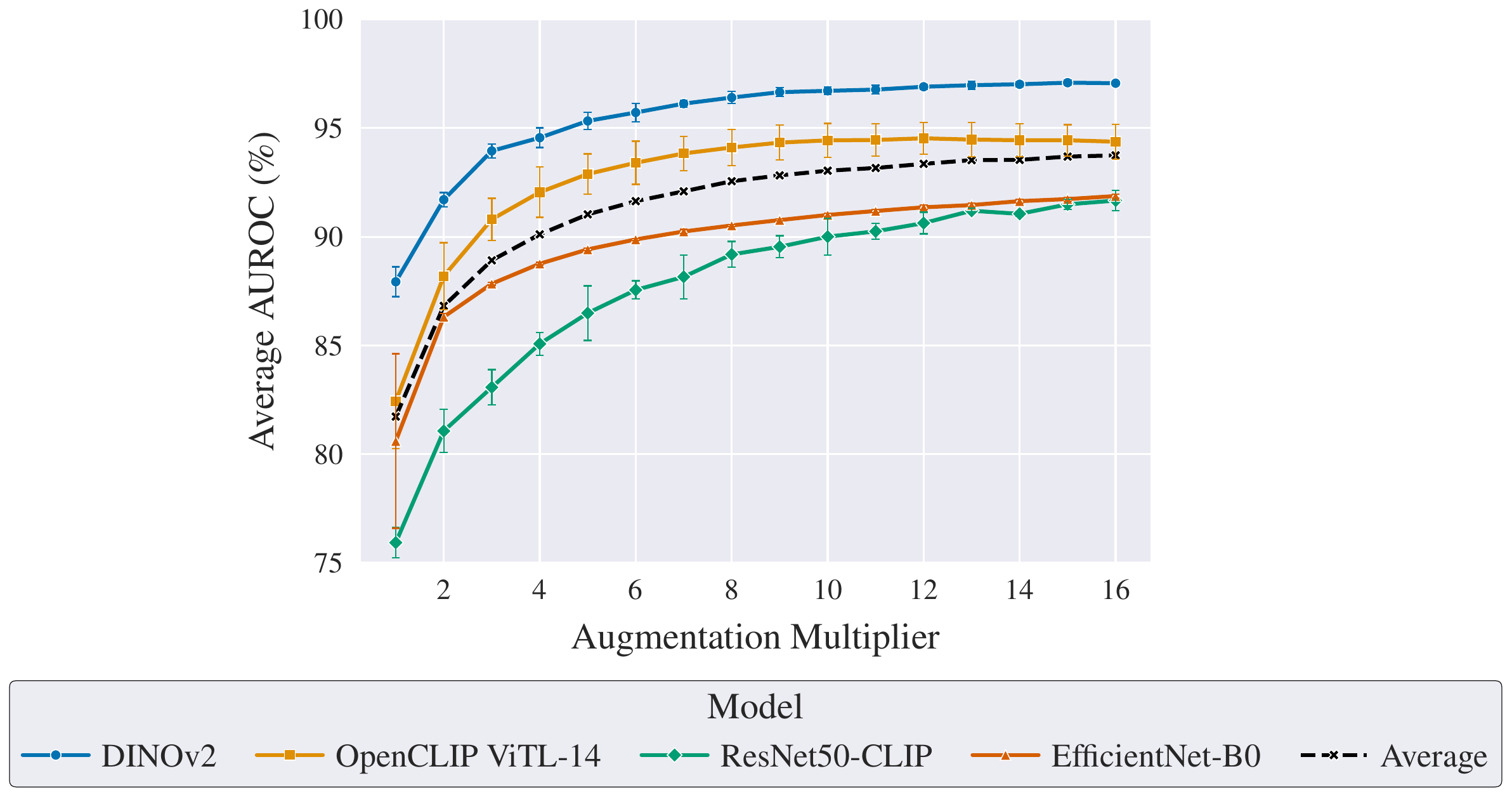}
\caption{Augmentation multiplier. Error bars denote the standard deviation across the three runs.}
\label{fig:am_variation}
\end{figure}

\begin{figure}[H]
\centering
\includegraphics[width=0.78\textwidth]{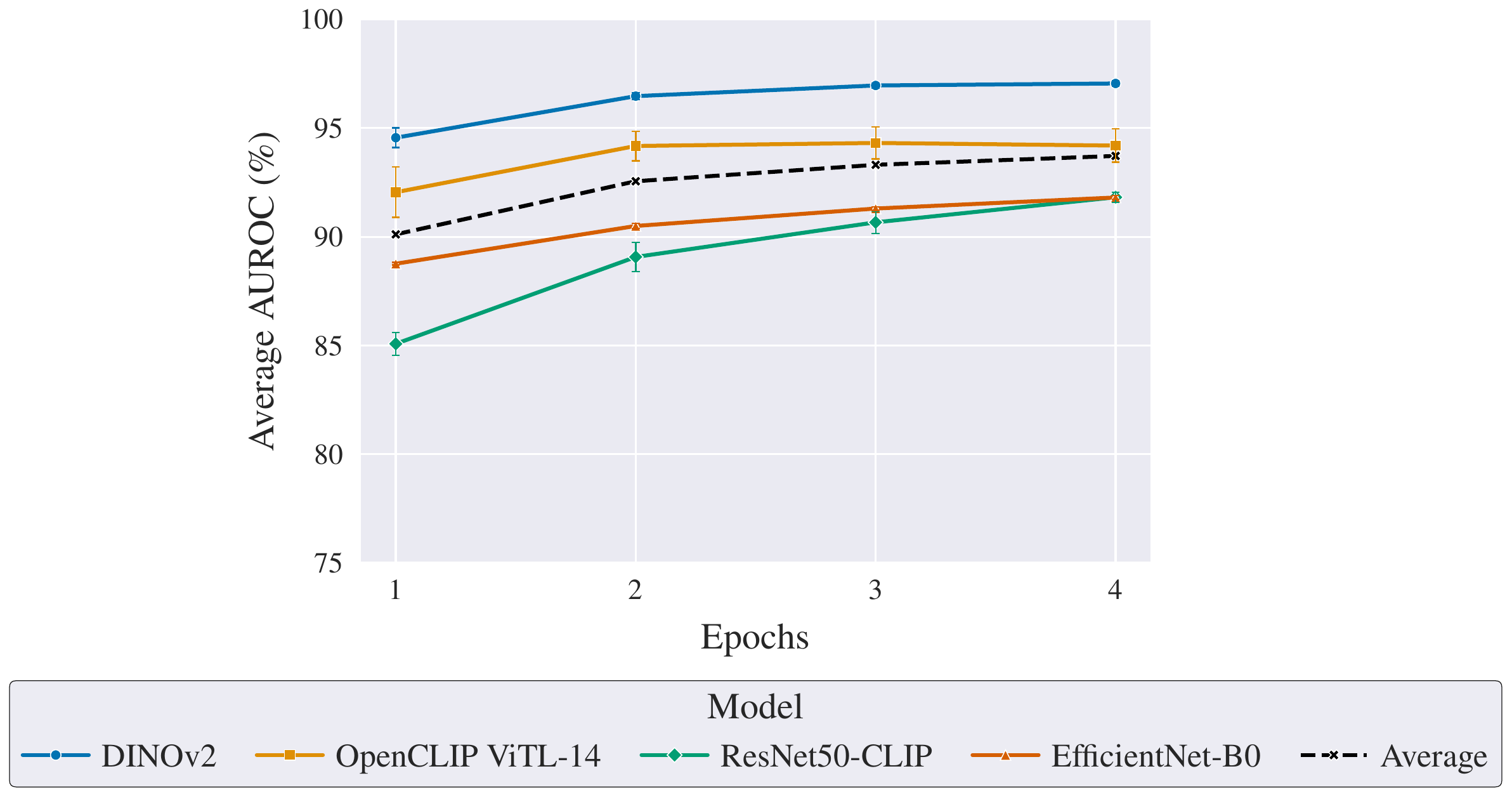}
\caption{Training duration. Error bars denote the standard deviation across the three runs.}
\label{fig:epochs_variation}
\end{figure}

Figures~\ref{fig:am_variation} and~\ref{fig:epochs_variation}, together with Tables~\ref{tab:am_variation} and~\ref{tab:epochs_variation}, summarize the results obtained by varying the augmentation multiplier ($am$) in the range $[1,16]$ while fixing the number of epochs to $1$, and by varying the number of epochs in $[1,4]$ while fixing $am=4$, as in the AI-GenBench paper.

\begin{table}[H]
\centering
\small
\setlength{\tabcolsep}{3.5pt}
\begin{tabular}{lccccc}
\toprule
\textbf{$am$} & \textbf{DINOv2} & \textbf{ViT-L CLIP} & \textbf{ResNet-50 CLIP} & \textbf{EfficientNet-B0} & \textbf{Average} \\
\midrule
1 & 87.93 $\pm$ 0.69 & 82.45 $\pm$ 2.18 & 75.94 $\pm$ 0.69 & 80.61 $\pm$ 4.01 & 81.73 \\
2 & 91.71 $\pm$ 0.33 & 88.19 $\pm$ 1.54 & 81.07 $\pm$ 1.00 & 86.32 $\pm$ 0.16 & 86.82 \\
3 & 93.95 $\pm$ 0.32 & 90.80 $\pm$ 0.97 & 83.08 $\pm$ 0.81 & 87.84 $\pm$ 0.07 & 88.92 \\
4 & 94.56 $\pm$ 0.45 & 92.05 $\pm$ 1.16 & 85.08 $\pm$ 0.52 & 88.76 $\pm$ 0.06 & 90.11 \\
5 & 95.33 $\pm$ 0.39 & 92.89 $\pm$ 0.93 & 86.49 $\pm$ 1.25 & 89.42 $\pm$ 0.07 & 91.03 \\
6 & 95.72 $\pm$ 0.42 & 93.40 $\pm$ 0.99 & 87.56 $\pm$ 0.42 & 89.88 $\pm$ 0.04 & 91.64 \\
7 & 96.12 $\pm$ 0.16 & 93.83 $\pm$ 0.78 & 88.16 $\pm$ 1.01 & 90.24 $\pm$ 0.09 & 92.09 \\
8 & 96.40 $\pm$ 0.27 & 94.11 $\pm$ 0.83 & 89.19 $\pm$ 0.59 & 90.52 $\pm$ 0.04 & 92.56 \\
9 & 96.65 $\pm$ 0.20 & 94.33 $\pm$ 0.80 & 89.54 $\pm$ 0.51 & 90.77 $\pm$ 0.04 & 92.82 \\
10 & 96.71 $\pm$ 0.17 & 94.44 $\pm$ 0.78 & 90.01 $\pm$ 0.84 & 91.01 $\pm$ 0.08 & 93.04 \\
11 & 96.77 $\pm$ 0.20 & 94.45 $\pm$ 0.75 & 90.25 $\pm$ 0.37 & 91.19 $\pm$ 0.07 & 93.17 \\
12 & 96.90 $\pm$ 0.12 & \textbf{94.53 $\pm$ 0.73} & 90.63 $\pm$ 0.50 & 91.36 $\pm$ 0.10 & 93.35 \\
13 & 96.97 $\pm$ 0.19 & 94.47 $\pm$ 0.79 & 91.20 $\pm$ 0.12 & 91.46 $\pm$ 0.09 & 93.53 \\
14 & 97.01 $\pm$ 0.11 & 94.44 $\pm$ 0.77 & 91.06 $\pm$ 0.12 & 91.64 $\pm$ 0.09 & 93.54 \\
15 & \textbf{97.09 $\pm$ 0.11} & 94.44 $\pm$ 0.72 & 91.49 $\pm$ 0.24 & 91.74 $\pm$ 0.09 & 93.69 \\
16 & 97.06 $\pm$ 0.09 & 94.37 $\pm$ 0.80 & \textbf{91.67 $\pm$ 0.46} & \textbf{91.87 $\pm$ 0.08} & \textbf{93.74} \\
\bottomrule
\end{tabular}
\caption{Effect of the augmentation multiplier ($am$) on the Next Period AUROC averaged over all temporal evaluation steps (\%). Values are mean $\pm$ standard deviation over the three runs; the Average column reports the mean across backbones. The best result per backbone (highest mean) is highlighted in bold. Bootstrap 95\% confidence intervals are reported in Appendix~\ref{app:bootstrap}.}
\label{tab:am_variation}
\end{table}

\begin{table}[H]
\centering
\small
\setlength{\tabcolsep}{3.5pt}
\begin{tabular}{lccccc}
\toprule
\textbf{Epochs} & \textbf{DINOv2} & \textbf{ViT-L CLIP} & \textbf{ResNet-50 CLIP} & \textbf{EfficientNet-B0} & \textbf{Average} \\
\midrule
1 & 94.56 $\pm$ 0.45 & 92.05 $\pm$ 1.16 & 85.08 $\pm$ 0.52 & 88.76 $\pm$ 0.06 & 90.11 \\
2 & 96.47 $\pm$ 0.17 & 94.18 $\pm$ 0.68 & 89.07 $\pm$ 0.67 & 90.50 $\pm$ 0.12 & 92.55 \\
3 & 96.96 $\pm$ 0.12 & \textbf{94.32 $\pm$ 0.74} & 90.67 $\pm$ 0.52 & 91.30 $\pm$ 0.09 & 93.31 \\
4 & \textbf{97.05 $\pm$ 0.13} & 94.19 $\pm$ 0.76 & \textbf{91.82 $\pm$ 0.23} & \textbf{91.81 $\pm$ 0.08} & \textbf{93.72} \\
\bottomrule
\end{tabular}
\caption{Effect of the training duration (epochs) on the Next Period AUROC averaged over all temporal evaluation steps (\%). Values are mean $\pm$ standard deviation over the three runs; the Average column reports the mean across backbones. The best result per backbone (highest mean) is highlighted in bold. Bootstrap 95\% confidence intervals are reported in Appendix~\ref{app:bootstrap}.}
\label{tab:epochs_variation}
\end{table}

We observe that larger models, such as ViT-L CLIP and DINOv2, reach a performance plateau more quickly than smaller backbones like ResNet-50 CLIP and EfficientNet-B0. In particular, ResNet-50 CLIP continues to benefit from longer training schedules, whereas transformer-based models converge after only one or two epochs. EfficientNet-B0 reaches a good AUROC score even with just one epoch, surpassing ResNet-50 CLIP. However, with four epochs ResNet-50 CLIP catches up, and the two backbones become virtually indistinguishable ($91.82\%$ vs. $91.81\%$).
When comparing the effect of increasing $am$ versus increasing the number of epochs, the two strategies appear equivalent. For example, configurations \textit{$am=8$, epochs=1} and \textit{$am=4$, epochs=2} yield similar AUROC values ($96.40\%$ vs $96.47\%$ for DINOv2). This suggests that dataset expansion through augmentation and repeated exposure to the same augmented samples both provide sufficient ``fuel'' for training, with no clear advantage of one approach over the other.
Overall, these results indicate that, while the augmentation multiplier can effectively replace longer training schedules, small-capacity models may still benefit from additional epochs before reaching their performance ceiling. Finally, it is worth noting that, while increasing the augmentation multiplier $am$ could be a reasonable choice for practical deployments, the AI-GenBench fairness rules prohibit values above $am=4$ to constrain training data diversity and ensure comparability across approaches. In contrast, increasing the number of epochs is allowed.

\subsection{Input processing at training and inference time}
The AI-GenBench paper established a strong baseline where models are trained on resized images (downsized to the model’s input resolution) and evaluated in the same way. In their study, this resize-based setup outperformed an alternative configuration where models were trained on random crops and evaluated via multi-crop inference (with scores averaged across crops). Multi-crop inference, in turn, proved to be better than single (center)-crop inference.

\begin{figure}[H]
\centering
\includegraphics[width=0.78\textwidth]{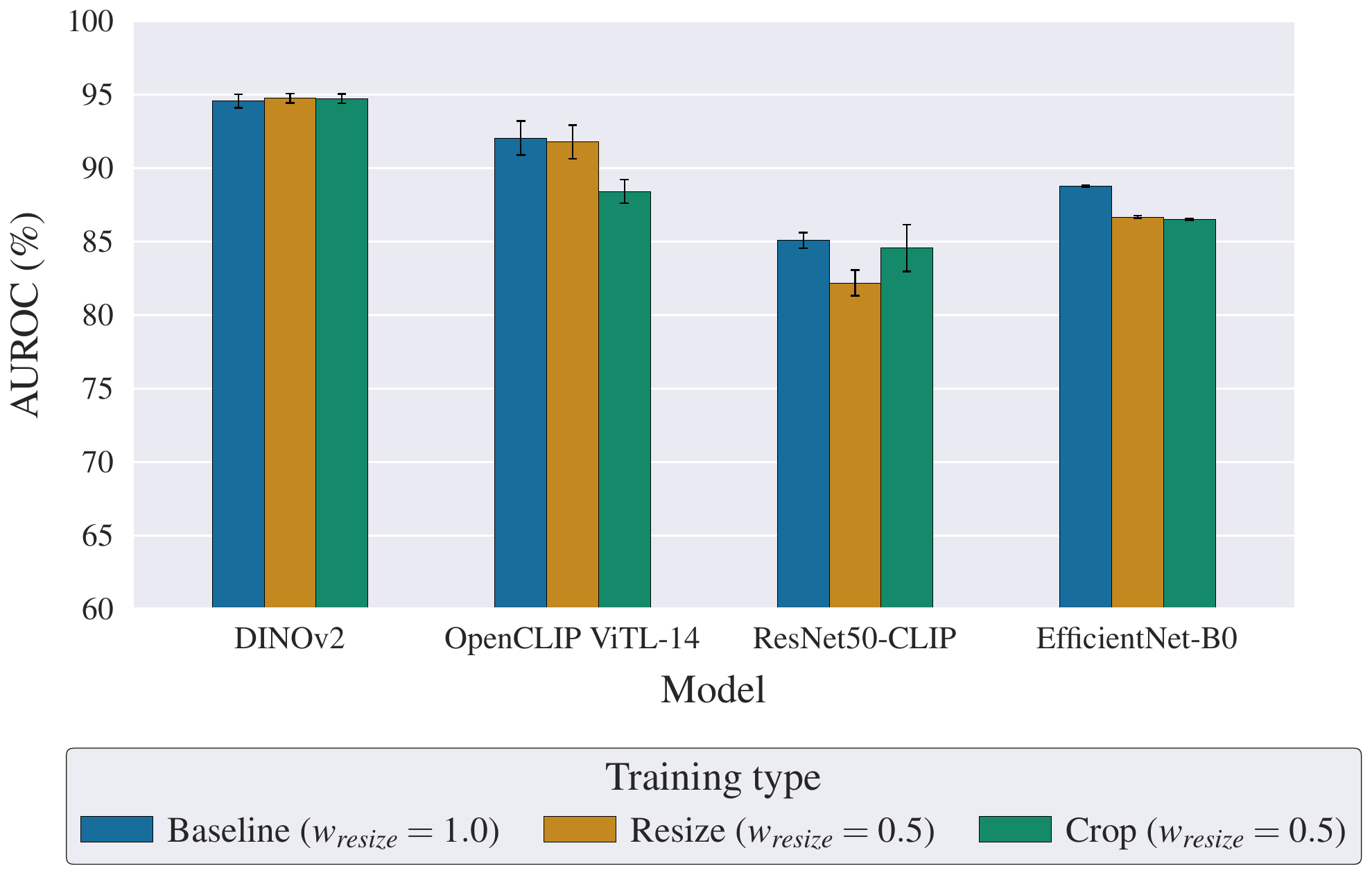}
\caption{Input processing. Error bars denote the standard deviation across the three runs.}
\label{fig:crop_resize_variation}
\end{figure}

In our work, we extend this comparison by introducing a hybrid evaluation strategy that combines both resized and cropped inputs. Specifically, at inference time, we generate five crops per image, average their prediction scores, and then combine this crop-based score with the resized-image score using equal weighting ($w=0.5$ for both). This \emph{Mixed} evaluation strategy is applied separately to both resize-trained and crop-trained models. Figure~\ref{fig:crop_resize_variation} and Table~\ref{tab:crop_resize_variation} summarize the results. The relative effectiveness of each strategy depends on the backbone architecture:
\begin{itemize}
    \item ResNet-50 CLIP - mixed evaluation does not provide improvements: the resize-only baseline clearly outperforms resize-trained models with mixed evaluation, while its advantage over crop-trained models with mixed evaluation ($85.08\%$ vs. $84.56\%$) is within the standard deviation across runs.
    \item ViT-L CLIP - crop-trained models underperform, while resize-trained models with mixed evaluation achieve results comparable to the resize-only baseline.
    \item DINOv2 - the three configurations are practically indistinguishable: both mixed evaluation variants (resize-trained, $94.75\%$, and crop-trained, $94.73\%$) are only nominally above the resize-only baseline ($94.56\%$), with differences well within the standard deviation across runs.
    \item EfficientNet-B0 - the resize-only baseline remains the best option, while adding crop-based scores reduces performance.
\end{itemize}

These findings suggest that the effect of the input processing strategy is architecture-dependent. For the larger backbones (DINOv2 and ViT-L CLIP), incorporating multi-crop information at inference time does not bring a measurable benefit over the resize-only baseline, whereas for smaller backbones like ResNet-50 CLIP and EfficientNet-B0 adding crop-based scores to resize-trained models clearly degrades performance (by $2.90$ and $2.09$ percentage points, respectively). Training and evaluating solely on resized images is therefore the most reliable choice across the evaluated architectures.

\begin{table}[H]
\centering
\small
\setlength{\tabcolsep}{2.5pt}
\begin{tabular}{llcccc}
\toprule
\textbf{Training input} & \textbf{Evaluation input} & \textbf{DINOv2} & \textbf{ViT-L CLIP} & \textbf{ResNet-50 CLIP} & \textbf{EfficientNet-B0} \\
\midrule
Resize & Resize & 94.56 $\pm$ 0.45 & \textbf{92.05 $\pm$ 1.16} & \textbf{85.08 $\pm$ 0.52} & \textbf{88.76 $\pm$ 0.06} \\
Resize & Mixed & \textbf{94.75 $\pm$ 0.31} & 91.77 $\pm$ 1.15 & 82.18 $\pm$ 0.88 & 86.67 $\pm$ 0.10 \\
Crop & Mixed & 94.73 $\pm$ 0.31 & 88.42 $\pm$ 0.82 & 84.56 $\pm$ 1.59 & 86.50 $\pm$ 0.08 \\
\bottomrule
\end{tabular}
\caption{Comparison of training and evaluation input strategies (Next Period AUROC averaged over all temporal evaluation steps, \%). Values are mean $\pm$ standard deviation over the three runs. \emph{Mixed} refers to the fusion of the scores obtained from the crops and from the resized version of the full image. The first row corresponds to the \emph{Baseline} model. The best result per backbone (highest mean) is highlighted in bold. Bootstrap 95\% confidence intervals are reported in Appendix~\ref{app:bootstrap}.}
\label{tab:crop_resize_variation}
\end{table}

\subsection{Multiclass training}
An open question in deepfake detection is whether exploiting generator labels during training can improve binary classification performance. While the task is typically formulated as a binary problem (real vs.~synthetic), it can also be reframed as a multiclass problem (one real class plus one class per generator). The key challenge then becomes how to map multiclass outputs to a single binary prediction at evaluation time. To investigate this, we evaluate three strategies: \textit{i)} plain multiclass training with fusion to binary, \textit{ii)} dual-head training combining binary and multiclass losses, and \textit{iii)} a distance-based approach using class centroids.

\subsubsection{Plain multiclass training}
\begin{figure}[H]
\centering
\includegraphics[width=0.78\textwidth]{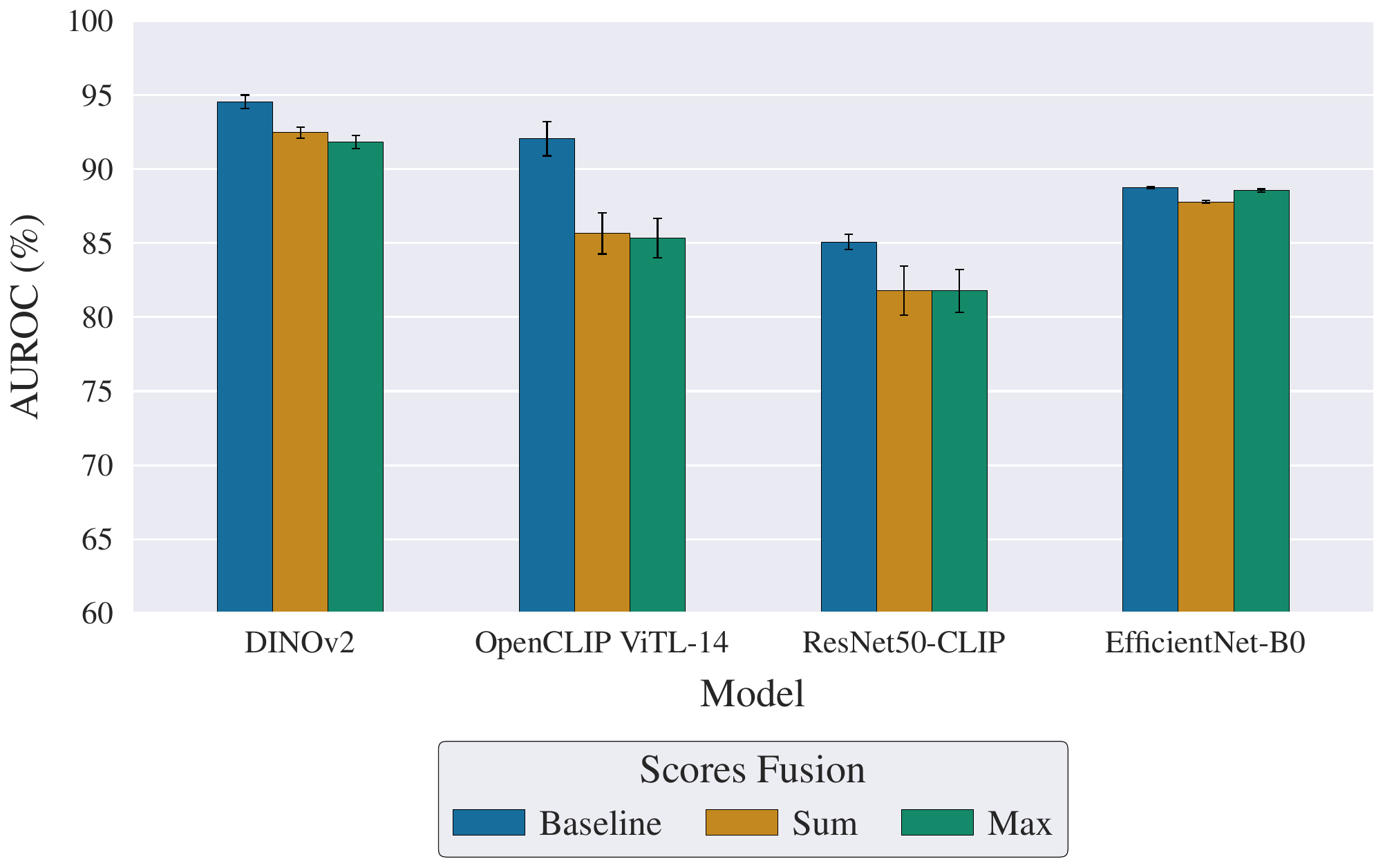}
\caption{Plain multiclass training. Error bars denote the standard deviation across the three runs.}
\label{fig:multiclass_variation}
\end{figure}

In the first setup, models are trained to predict generator identities directly. At inference time, the multiclass outputs are mapped to a binary decision using either \emph{Sum fusion} or \emph{Max fusion}. Figure~\ref{fig:multiclass_variation} and Table~\ref{tab:multiclass_plain} report the results. Direct binary training (the \emph{baseline}) outperforms plain multiclass training followed by fusion across all models. The gap is especially pronounced for the ViT-L CLIP backbone, where binary training yields a significantly higher AUROC score ($92.05\%$ vs. $85.65\%$ and $85.35\%$ with sum and max fusion, respectively). For DINOv2, ResNet-50 CLIP, and EfficientNet-B0, the difference is smaller; in particular, EfficientNet-B0 with max fusion nearly matches the binary baseline. When comparing fusion strategies, the two are essentially on par on average, and the only difference that clearly exceeds the variability across runs is observed for EfficientNet-B0, where max fusion outperforms sum fusion ($88.54\%$ vs. $87.78\%$); in any case, neither closes the gap with binary training. These results suggest that while generator-specific supervision encourages richer representations, simply collapsing them into a binary decision at inference time is less effective than directly optimizing for the binary objective.

\begin{table}[H]
\centering
\small
\setlength{\tabcolsep}{3.5pt}
\begin{tabular}{lcccc}
\toprule
\textbf{Fusion strategy} & \textbf{DINOv2} & \textbf{ViT-L CLIP} & \textbf{ResNet-50 CLIP} & \textbf{EfficientNet-B0} \\
\midrule
Baseline & \textbf{94.56 $\pm$ 0.45} & \textbf{92.05 $\pm$ 1.16} & \textbf{85.08 $\pm$ 0.52} & \textbf{88.76 $\pm$ 0.06} \\
Sum fusion & 92.47 $\pm$ 0.37 & 85.65 $\pm$ 1.38 & 81.79 $\pm$ 1.64 & 87.78 $\pm$ 0.10 \\
Max fusion & 91.84 $\pm$ 0.44 & 85.35 $\pm$ 1.33 & 81.77 $\pm$ 1.44 & 88.54 $\pm$ 0.11 \\
\bottomrule
\end{tabular}
\caption{Plain multiclass training: comparison of fusion strategies (Next Period AUROC averaged over all temporal evaluation steps, \%). Values are mean $\pm$ standard deviation over the three runs. \emph{Baseline} refers to direct binary training. The best result per backbone (highest mean) is highlighted in bold. Bootstrap 95\% confidence intervals are reported in Appendix~\ref{app:bootstrap}.}
\label{tab:multiclass_plain}
\end{table}

\subsubsection{Dual-head training}
Starting from the previous observation, we evaluated whether combining binary and multiclass supervision via a dual-head architecture can improve detection performance. For each configuration (\emph{Separate heads} and \emph{Stacked heads}), we explored two loss-weighting schemes: \textit{i)} equal weighting ($w_{\text{bin}} = w_{\text{multi}} = 0.5$), and \textit{ii)} auxiliary weighting, where the binary loss dominates ($w_{\text{bin}} = 0.75$, $w_{\text{multi}} = 0.25$) treating the multiclass signal as an auxiliary objective.

\begin{figure}[H]
\centering
\includegraphics[width=0.78\textwidth]{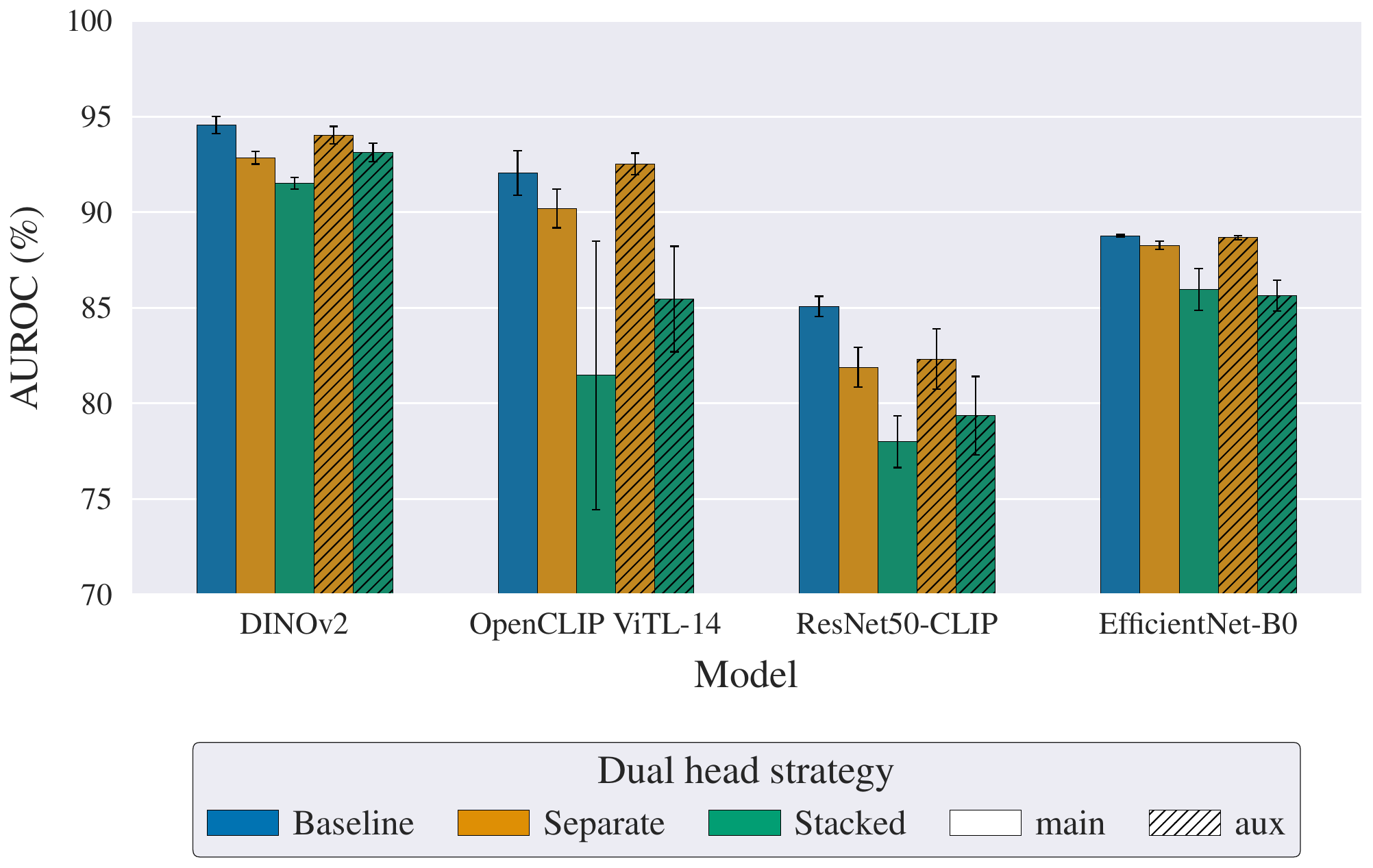}
\caption{Dual-head training. Error bars denote the standard deviation across the three runs.}
\label{fig:multihead_variation}
\end{figure}

Figure~\ref{fig:multihead_variation} and Table~\ref{tab:multihead} report the results of the four configurations and the baseline. Several consistent patterns emerge:
\begin{itemize}
    \item The \emph{Separate heads} approach consistently outperforms the \emph{Stacked heads} approach across all backbones.
    \item Using the multiclass loss as an auxiliary signal (aux) is generally superior to equal weighting, and the best dual-head result is obtained by separate heads with auxiliary weighting for each backbone.
    \item Among the four dual-head combinations, separate heads with auxiliary weighting achieves the best performance.
\end{itemize}

When compared to the pure binary baseline, this best dual-head strategy is nominally superior for ViT-L CLIP ($92.53\%$ vs. $92.05\%$) and slightly inferior for DINOv2 ($94.02\%$ vs. $94.56\%$), in both cases by about one standard deviation across runs; it is also slightly inferior for EfficientNet-B0 ($88.67\%$ vs. $88.76\%$), while it causes a consistent drop for ResNet-50 CLIP ($82.32\%$ vs. $85.08\%$). These results suggest that employing an auxiliary supervision based on the generator label preserves the detection performance of the larger backbones, without measurably improving it, whereas it is detrimental for ResNet-50 CLIP. Fine-tuning the relative loss weights may yield further improvements, but this is beyond the scope of the present study.

\begin{table}[H]
\centering
\small
\setlength{\tabcolsep}{3.5pt}
\begin{tabular}{lcccc}
\toprule
\textbf{Configuration} & \textbf{DINOv2} & \textbf{ViT-L CLIP} & \textbf{ResNet-50 CLIP} & \textbf{EfficientNet-B0} \\
\midrule
Baseline & \textbf{94.56 $\pm$ 0.45} & 92.05 $\pm$ 1.16 & \textbf{85.08 $\pm$ 0.52} & \textbf{88.76 $\pm$ 0.06} \\
Separate heads & 92.84 $\pm$ 0.33 & 90.20 $\pm$ 1.01 & 81.89 $\pm$ 1.05 & 88.27 $\pm$ 0.21 \\
Stacked heads & 91.52 $\pm$ 0.30 & 81.47 $\pm$ 7.02 & 78.00 $\pm$ 1.35 & 85.96 $\pm$ 1.10 \\
Separate heads (aux) & 94.02 $\pm$ 0.46 & \textbf{92.53 $\pm$ 0.56} & 82.32 $\pm$ 1.57 & 88.67 $\pm$ 0.10 \\
Stacked heads (aux) & 93.11 $\pm$ 0.48 & 85.46 $\pm$ 2.76 & 79.36 $\pm$ 2.05 & 85.64 $\pm$ 0.80 \\
\bottomrule
\end{tabular}
\caption{Dual-head training results (Next Period AUROC averaged over all temporal evaluation steps, \%). Values are mean $\pm$ standard deviation over the three runs. \emph{Aux} denotes down-weighting the multiclass loss ($w_{bin}=0.75, w_{multi}=0.25$). \emph{Baseline} refers to training using a binary loss only. The best result per backbone (highest mean) is highlighted in bold. Bootstrap 95\% confidence intervals are reported in Appendix~\ref{app:bootstrap}.}
\label{tab:multihead}
\end{table}

\subsubsection{MLP vs distance-based approach}
\begin{figure}[H]
\centering
\includegraphics[width=0.78\textwidth]{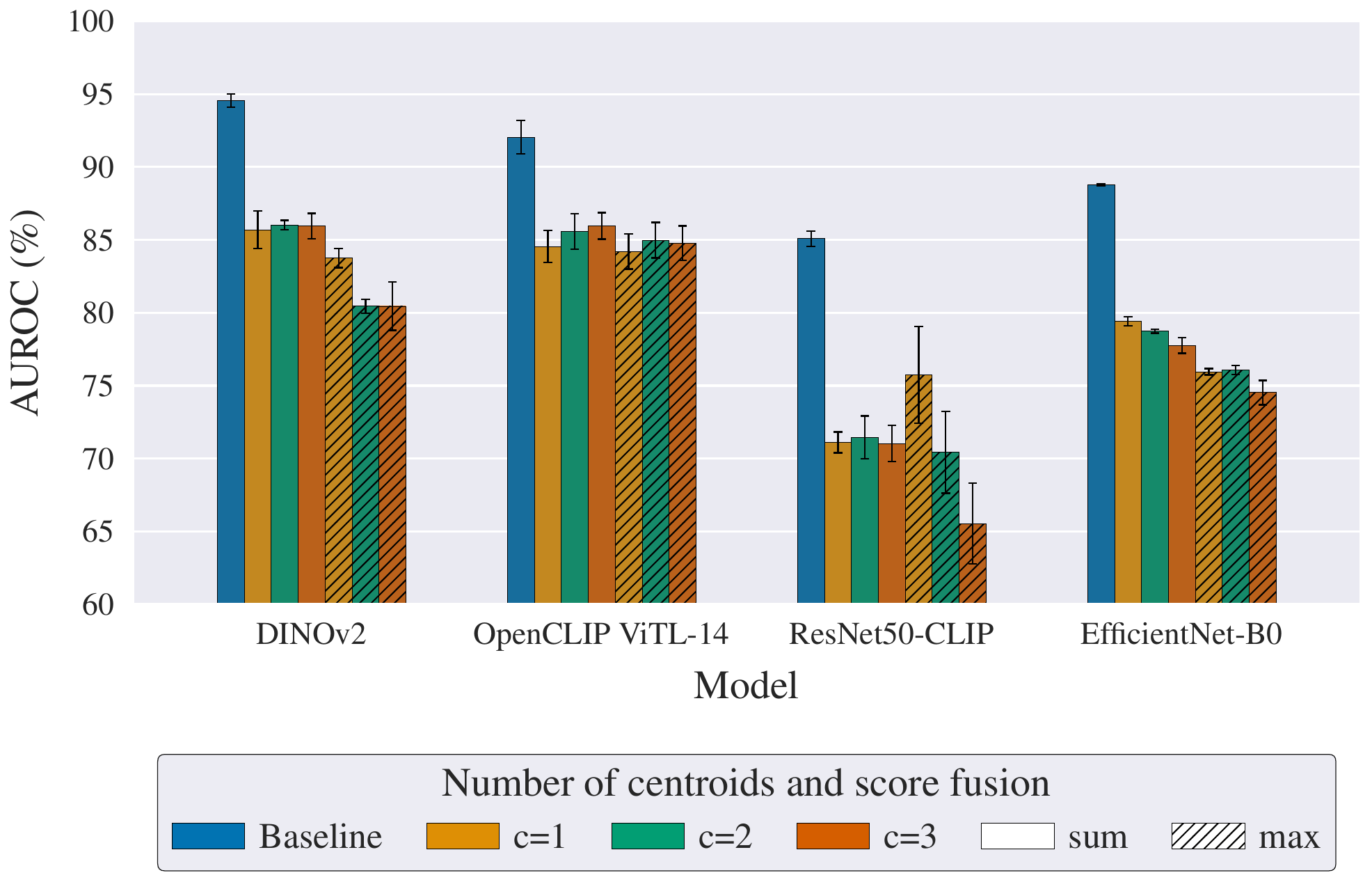}
\caption{Distance-based scoring. Error bars denote the standard deviation across the three runs.}
\label{fig:centroid_variation}
\end{figure}

Figure~\ref{fig:centroid_variation} and Table~\ref{tab:centroid_variation} summarize the results for the distance-based approach. Across all backbones, the \emph{baseline} MLP trained directly on the binary task remains superior. Compared to the plain multiclass setup with fusion (see Table~\ref{tab:multiclass_plain}), even the best centroid configuration underperforms on DINOv2, ResNet-50 CLIP, and EfficientNet-B0, while for ViT-L CLIP it is on par with the plain multiclass approach ($85.97\%$ vs. $85.65\%$, a difference within the standard deviation across runs). In all cases, it still falls notably short of the binary baseline.

Regarding fusion strategies within the distance-based approach, \emph{sum fusion} tends to outperform \emph{max fusion} on the larger backbones (DINOv2 and ViT-L CLIP) and on EfficientNet-B0, whereas for ResNet-50 CLIP \emph{max fusion} performs better only with a single centroid ($75.74\%$ vs. $71.10\%$, with a large standard deviation across runs) and \emph{sum fusion} prevails for $c=2$ and $c=3$ (in all cases, far below the MLP baseline). Overall, these mixed outcomes indicate that centroid-based scoring is generally less effective and less reliable than a standard MLP head trained on the binary task.

\begin{table}[H]
\centering
\small
\setlength{\tabcolsep}{3.5pt}
\begin{tabular}{llcccc}
\toprule
\textbf{Centroids} & \textbf{Fusion} & \textbf{DINOv2} & \textbf{ViT-L CLIP} & \textbf{ResNet-50 CLIP} & \textbf{EfficientNet-B0} \\
\midrule
Baseline & -- & \textbf{94.56 $\pm$ 0.45} & \textbf{92.05 $\pm$ 1.16} & \textbf{85.08 $\pm$ 0.52} & \textbf{88.76 $\pm$ 0.06} \\
1 & Sum & 85.69 $\pm$ 1.29 & 84.55 $\pm$ 1.09 & 71.10 $\pm$ 0.72 & 79.42 $\pm$ 0.31 \\
2 & Sum & 86.01 $\pm$ 0.32 & 85.59 $\pm$ 1.22 & 71.46 $\pm$ 1.47 & 78.73 $\pm$ 0.13 \\
3 & Sum & 85.95 $\pm$ 0.86 & 85.97 $\pm$ 0.91 & 71.04 $\pm$ 1.23 & 77.76 $\pm$ 0.54 \\
1 & Max & 83.75 $\pm$ 0.66 & 84.20 $\pm$ 1.20 & 75.74 $\pm$ 3.32 & 75.95 $\pm$ 0.21 \\
2 & Max & 80.45 $\pm$ 0.48 & 84.98 $\pm$ 1.22 & 70.43 $\pm$ 2.82 & 76.07 $\pm$ 0.32 \\
3 & Max & 80.45 $\pm$ 1.65 & 84.78 $\pm$ 1.18 & 65.53 $\pm$ 2.77 & 74.53 $\pm$ 0.84 \\
\bottomrule
\end{tabular}
\caption{Distance-based scoring vs.\ MLP baseline (Next Period AUROC averaged over all temporal evaluation steps, \%). Values are mean $\pm$ standard deviation over the three runs. The best result per backbone (highest mean) is highlighted in bold. Bootstrap 95\% confidence intervals are reported in Appendix~\ref{app:bootstrap}.}
\label{tab:centroid_variation}
\end{table}

\section{Incremental Update Strategy}
\label{sec:ius}
An additional dimension in our study concerns how detectors can be efficiently and effectively updated as new generators are released over time. In the AI-GenBench evaluation framework, this corresponds to progressing through successive temporal windows, when (four) new generators are introduced at each step.

\subsubsection*{Baseline (batch tuning)} 
The standard setting, introduced in the AI-GenBench paper \cite{pellegrini2025ai}, consists of successive \emph{batch tuning} steps on all data accumulated up to the current window, starting from the model weights obtained up to that moment. This strategy has two consequences: \textit{i)} the model may benefit from the fact that past generators were already learned, and \textit{ii)} past and new generators are equally represented in the cumulative training set. While effective, this approach is computationally expensive, as it requires retraining on the entire generator history at each step.

\subsubsection*{Reset weights}
As a reference, we consider a variant in which, at each window, the detector is retrained on the cumulative data but initialized from the original pretrained weights rather than from the model obtained at the previous step. This design removes the bias in favor of earlier generators present in the baseline approach but discards previously consolidated knowledge that could be beneficial when adapting to new generators. In our experiments, we compare this strategy against both the baseline and continual learning approaches.

\subsection{Continual learning strategies}
We explore several continual learning strategies aimed at balancing adaptability to new generators with retention of knowledge about older ones. These strategies are designed to be more efficient than the batch retraining baseline while mitigating catastrophic forgetting. In particular, we focus on replay-based strategies, which reduce forgetting by storing and replaying a subset of images from generators encountered in previous windows. We evaluate the performance of both a size-unbounded and a size-bounded replay strategy.

\subsubsection*{Harmonic replay}
In this strategy the number of stored samples per generator decreases according to a harmonic schedule. Initially, all training samples for each generator are inserted into the replay buffer. Over time, the contribution of each generator is reduced by a factor of $1/i$, where $i$ is the number of windows elapsed since that generator was introduced. This allocation reserves more space for recent generators while gradually down-weighting older ones. We refer to this as a \emph{Harmonic} schedule since the total buffer size grows without bound, following the harmonic series $N \cdot (1 + \tfrac{1}{2} + \tfrac{1}{3} + \dots)$.

\subsubsection*{Class-balanced replay}
Here, we consider a bounded replay buffer of fixed size, maintained using a class-balanced strategy in which an equal number of examples is kept for all generators. This means that as new generators are introduced, the portion of the replay buffer allocated to earlier generators decreases. This strategy enforces a strict memory budget, making the computational cost directly proportional to the size of the buffer. To assess how performance scales with replay capacity, we evaluate different buffer sizes (e.g., 10000 and 20000 samples).

\subsubsection*{Naive continual learning} 
Finally, as a lower bound, we evaluate a naive continual tuning approach where the model is initialized with the weights from the previous step but fine-tuned only on data from generators in the current window, without any replay or regularization mechanism.

\subsection{Results}

\begin{figure}[H]
\centering
\includegraphics[width=0.78\textwidth]{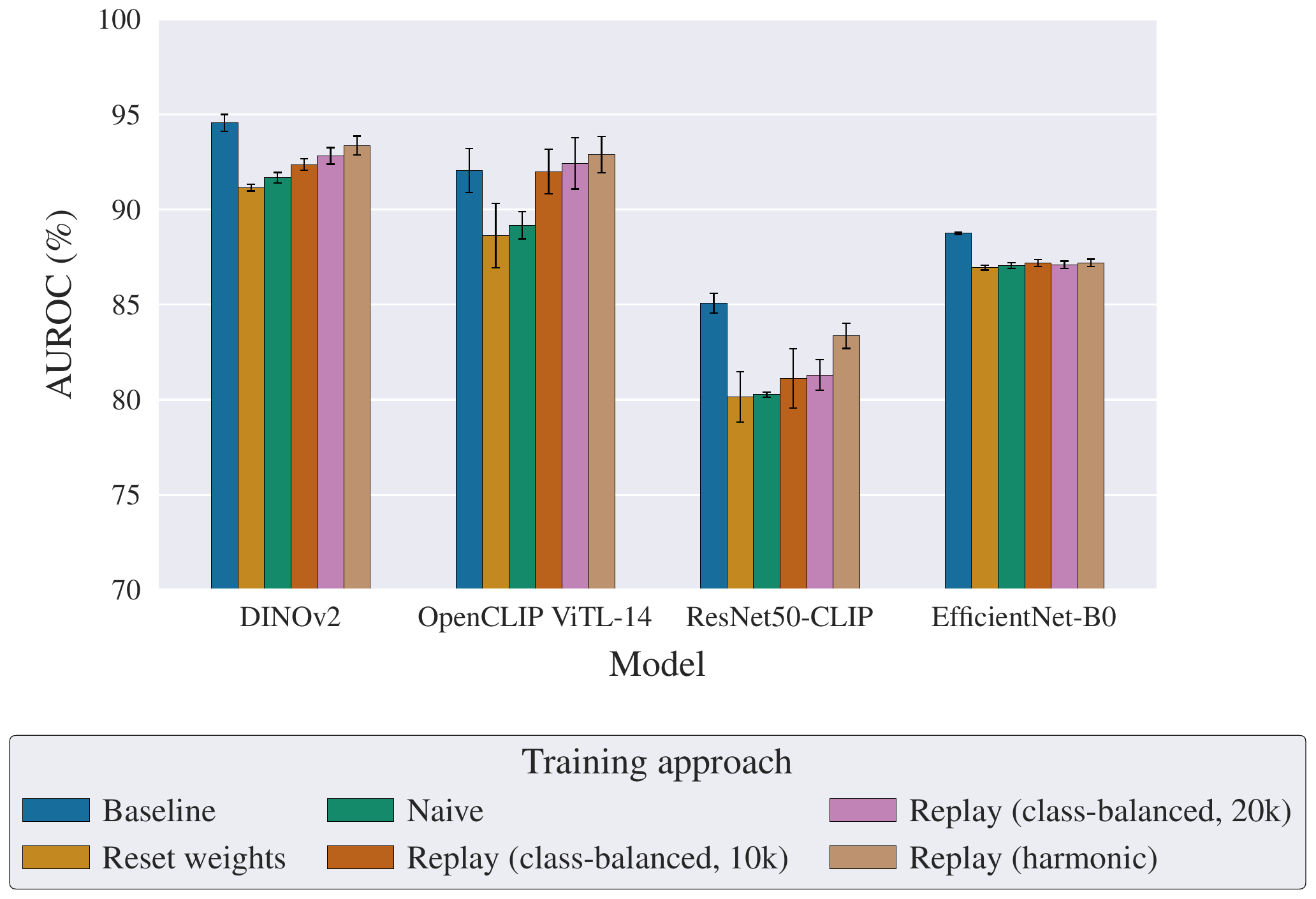}
\caption{Continual learning, Next Period. Error bars denote the standard deviation across the three runs.}
\label{fig:cl_next}
\end{figure}

\begin{figure}[H]
\centering
\includegraphics[width=0.78\textwidth]{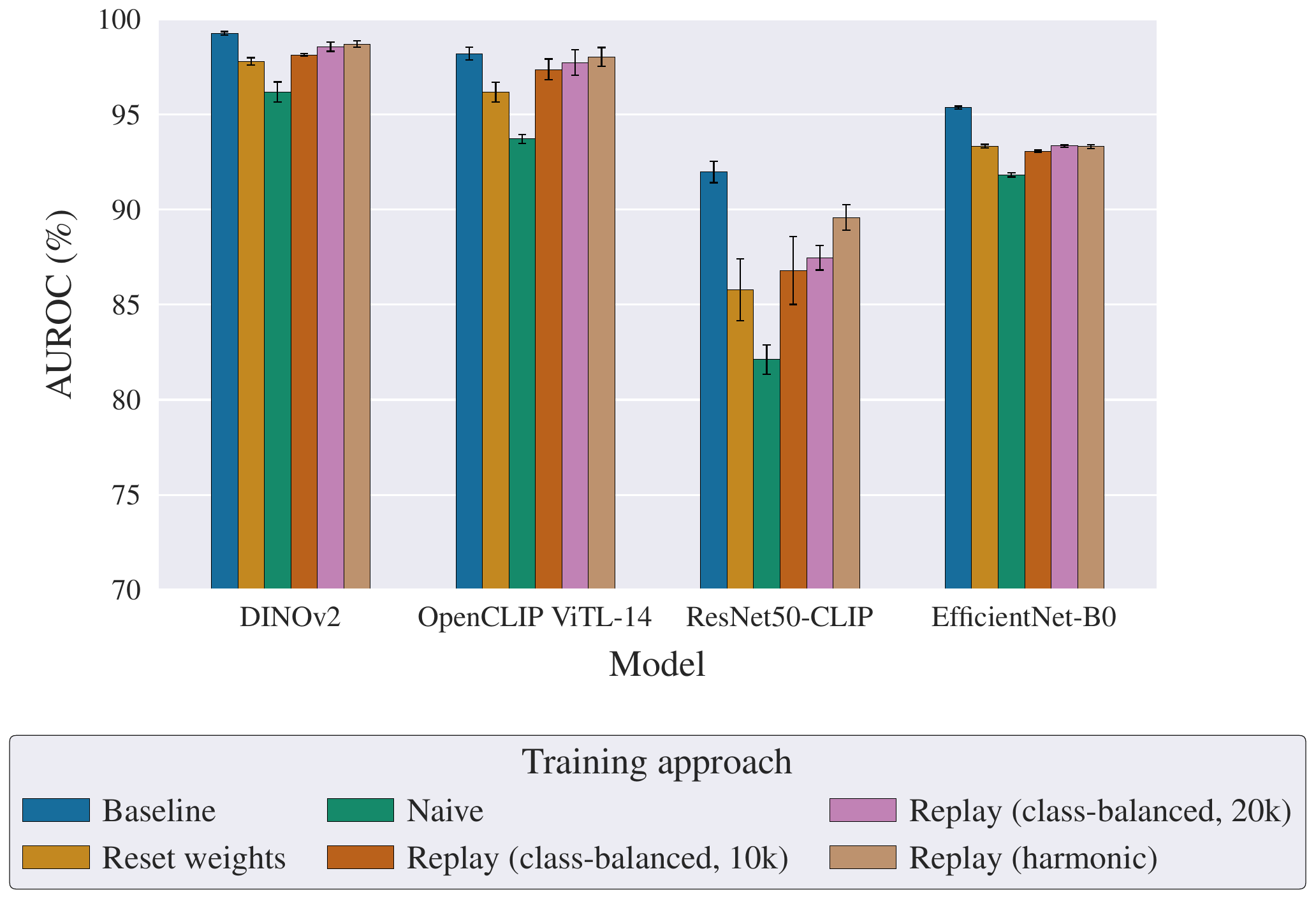}
\caption{Continual learning, Past Period. Error bars denote the standard deviation across the three runs.}
\label{fig:cl_past}
\end{figure}

Results for Next Period and Past Period AUROC are summarized in Figures~\ref{fig:cl_next} and~\ref{fig:cl_past}, with corresponding numerical values reported in Tables~\ref{tab:cl_next} and~\ref{tab:cl_past}. As expected, the naive approach preserves adaptability to new generators but suffers from severe forgetting on older ones. Replay buffers substantially mitigate this effect, with both class-balanced and harmonic replay offering a favorable trade-off between buffer size and retention. Resetting the model weights generally performs worse than both the batch baseline and replay-based approaches, confirming that using previous weights is beneficial; the only exception is EfficientNet-B0 on the \textit{Past Period}, where it is on par with the replay variants. Overall, continual learning with replay effectively balances adaptability and retention, while naive tuning alone is insufficient. For EfficientNet-B0, the performance of all the approaches is very close: on the \textit{Next Period}, harmonic replay obtains the highest mean, but the three replay variants differ by only $0.10$ percentage points, well within the standard deviation across runs. Forgetting, as measured by the performance gap between the batch baseline and the continual learning approaches on the \textit{Past Period}, is more pronounced for the smaller ResNet-50 CLIP model. This observation is consistent with recent findings in the literature showing that larger models, both in vision and language, exhibit greater resistance to forgetting \cite{ramasesh2022effect}.

\begin{table}[H]
\centering
\small
\setlength{\tabcolsep}{3.5pt}
\begin{tabular}{lcccc}
\toprule
\textbf{Configuration} & \textbf{DINOv2} & \textbf{ViT-L CLIP} & \textbf{ResNet-50 CLIP} & \textbf{EfficientNet-B0} \\
\midrule
Baseline (reference) & 94.56 $\pm$ 0.45 & 92.05 $\pm$ 1.16 & 85.08 $\pm$ 0.52 & 88.76 $\pm$ 0.06 \\
Reset weights & 91.16 $\pm$ 0.18 & 88.63 $\pm$ 1.69 & 80.14 $\pm$ 1.33 & 86.95 $\pm$ 0.13 \\
Naive & 91.68 $\pm$ 0.27 & 89.18 $\pm$ 0.72 & 80.28 $\pm$ 0.14 & 87.06 $\pm$ 0.15 \\
Replay (CB, 10k) & 92.37 $\pm$ 0.29 & 92.00 $\pm$ 1.17 & 81.12 $\pm$ 1.56 & 87.18 $\pm$ 0.19 \\
Replay (CB, 20k) & 92.82 $\pm$ 0.43 & 92.44 $\pm$ 1.35 & 81.31 $\pm$ 0.81 & 87.10 $\pm$ 0.19 \\
Replay (harmonic) & \textbf{93.37 $\pm$ 0.49} & \textbf{92.89 $\pm$ 0.95} & \textbf{83.36 $\pm$ 0.67} & \textbf{87.20 $\pm$ 0.19} \\
\bottomrule
\end{tabular}
\caption{Next Period performance of different training regimes (AUROC averaged over all temporal evaluation steps, \%). Values are mean $\pm$ standard deviation over the three runs. The best continual learning result per backbone (highest mean) is highlighted in bold. CB stands for \textit{class-balanced}, followed by the number of samples in the replay buffer. Bootstrap 95\% confidence intervals are reported in Appendix~\ref{app:bootstrap}.}
\label{tab:cl_next}
\end{table}

Results in Table~\ref{tab:cl_past} and Figure~\ref{fig:cl_past} confirm that both harmonic and class-balanced replay strategies greatly mitigate forgetting on generators introduced in past windows. Table~\ref{tab:cl_computation} shows the computational impact of these approaches: replay strategies significantly reduce the computational cost of adapting models to new generators compared to full batch retraining. In addition, it should be noted that the harmonic strategy, which shows the best continual learning results on all backbones for the \textit{Next Period} and on DINOv2, ViT-L CLIP, and ResNet-50 CLIP for the \textit{Past Period}, reduces the number of stored samples of generators over time, thus reducing their relevance at training time: generators become obsolete and content generated with those models is more easily recognizable. For EfficientNet-B0, the replay configurations obtain very similar performance (within $0.3$ percentage points on the \textit{Past Period}), making the differences among them negligible in practice. In addition, \textit{Past Period} performance shows that detection capabilities for older generators can be retained even with very few examples. Keeping these considerations in mind, decreasing the influence of older generators during training is both practical and natural.


\begin{table}[H]
\centering
\small
\setlength{\tabcolsep}{3.5pt}
\begin{tabular}{lcccc}
\toprule
\textbf{Configuration} & \textbf{DINOv2} & \textbf{ViT-L CLIP} & \textbf{ResNet-50 CLIP} & \textbf{EfficientNet-B0} \\
\midrule
Baseline (reference) & 99.27 $\pm$ 0.09 & 98.20 $\pm$ 0.34 & 91.97 $\pm$ 0.56 & 95.36 $\pm$ 0.08 \\
Reset weights & 97.80 $\pm$ 0.19 & 96.17 $\pm$ 0.51 & 85.78 $\pm$ 1.62 & 93.34 $\pm$ 0.09 \\
Naive & 96.19 $\pm$ 0.52 & 93.72 $\pm$ 0.23 & 82.12 $\pm$ 0.77 & 91.83 $\pm$ 0.11 \\
Replay (CB, 10k) & 98.14 $\pm$ 0.06 & 97.37 $\pm$ 0.55 & 86.79 $\pm$ 1.78 & 93.07 $\pm$ 0.06 \\
Replay (CB, 20k) & 98.57 $\pm$ 0.24 & 97.73 $\pm$ 0.68 & 87.47 $\pm$ 0.65 & \textbf{93.35 $\pm$ 0.07} \\
Replay (harmonic) & \textbf{98.71 $\pm$ 0.16} & \textbf{98.03 $\pm$ 0.49} & \textbf{89.59 $\pm$ 0.66} & 93.32 $\pm$ 0.10 \\
\bottomrule
\end{tabular}
\caption{Past Period performance of different training regimes (AUROC averaged over all temporal evaluation steps, \%). Values are mean $\pm$ standard deviation over the three runs. The best continual learning result per backbone (highest mean) is highlighted in bold. CB stands for \textit{class-balanced}, followed by the number of samples in the replay buffer. Bootstrap 95\% confidence intervals are reported in Appendix~\ref{app:bootstrap}.}
\label{tab:cl_past}
\end{table}

\section{Integrating the ``best of''}
\label{sec:best_of}
The main objective of this study was to identify which design practices transfer across different architectures and which should be selected according to the backbone. By systematically evaluating the impact of individual training and inference-time choices, we aimed to isolate configuration elements that lead to the most transferable improvements.

Excluding continual learning experiments, which address a distinct adaptation scenario, our results highlight a configuration that achieves the most consistent performance across the evaluated architectures. Specifically, while maintaining the augmentation multiplier at $am = 4$ in accordance with the AI-GenBench protocol, the best results were obtained with an extended training regimen of four epochs, both on average and for three of the four backbones (ViT-L CLIP peaks at three epochs, with a difference within the standard deviation across runs).

\begin{table}[t]
\centering
\setlength{\tabcolsep}{3.5pt}

\begin{tabular}{lccccc}
\toprule
\textbf{Window} & \textbf{Baseline} & \textbf{Naive} & \textbf{CB (10k)} & \textbf{CB (20k)} & \textbf{Harmonic} \\
\midrule
1 & 100\% & 50.00\% & 65.63\% & 75.00\% & 75.00\% \\
2 & 100\% & 33.33\% & 43.75\% & 54.17\% & 58.33\% \\
3 & 100\% & 25.00\% & 32.81\% & 40.63\% & 47.92\% \\
4 & 100\% & 20.00\% & 26.25\% & 32.50\% & 40.83\% \\
5 & 100\% & 16.67\% & 21.88\% & 27.08\% & 35.69\% \\
6 & 100\% & 14.29\% & 18.75\% & 23.21\% & 31.79\% \\
7 & 100\% & 12.50\% & 16.41\% & 20.31\% & 28.71\% \\
8 & 100\% & 11.11\% & 14.58\% & 18.06\% & 26.21\% \\
\midrule
\textbf{Average} & 100\% & 22.86\% & 30.01\% & 36.37\% & 43.06\% \\
\midrule
\ldots & \ldots & \ldots & \ldots & \ldots & \ldots \\ 
24 (20 years) & 100\% & 4.00\% & 5.25\% & 6.50\% & 11.55\% \\
\bottomrule
\end{tabular}
\caption{Relative computational cost per window as a percentage of the baseline. The setup uses AI-GenBench (32k new examples/window, divided into 16k real and 16k fake), with window 24 simulating a 20-year benchmark. CB stands for class-balanced, followed by the number of samples in the replay buffer.}
\label{tab:cl_computation}
\end{table}

Among the tested augmentation pipelines, the \emph{Evaluation-based} pipeline, featuring three probabilistic JPEG compression passes, proved to be the most effective. For input pre-processing, resizing the entire image to the model's input resolution emerged as the most reliable strategy across architectures, both during training and evaluation. Consequently, this approach was adopted for training the ``best of'' model. However, it should be noted that a hybrid strategy that combines full (resized) image training with prediction based on both the full image and a set of crops (five in our experiments) achieved comparable performance on larger models (DINOv2, VIT-L CLIP), but it does not provide a measurable gain ($+0.19$ and $-0.27$ percentage points, respectively, both within the standard deviation across runs) and increases inference cost because multiple forward passes are required.

Direct optimization of the binary classification objective remains the most reliable approach. However, on the larger backbones (DINOv2 and ViT-L CLIP), a dual-head configuration with an auxiliary head (and loss) jointly optimized on the multiclass generator labels achieves comparable results, within about one standard deviation, and offers the additional benefit of enabling \textit{model attribution}; on ResNet-50 CLIP, instead, it causes a consistent performance drop.

When these optimal design choices were combined and applied to the \textit{DINOv2} backbone, the highest-performing architecture among the evaluated models, the resulting configuration achieved an average AUROC of $97.36 \%$ on the \textit{Next Period} in a single training run, establishing the current state-of-the-art on AI-GenBench.

Table~\ref{tab:uncertainty_cost} summarizes the practical trade-off behind the main recommendations. The AUROC differences shown here are computed from the three-run means reported in the previous tables, while the 95\% bootstrap confidence intervals of the same differences are reported in Appendix~\ref{app:bootstrap}.

\begin{table}[H]
\centering
\footnotesize
\renewcommand{\arraystretch}{0.95}
\setlength{\tabcolsep}{3pt}
\begin{tabular}{p{3.6cm} l c c c}
\toprule
\textbf{Recommendation} &
\textbf{Backbone} &
\textbf{AUROC $\Delta$ (pp)} &
\textbf{Train} &
\textbf{Infer.} \\
\midrule

Evaluation augmentation \newline
vs. baseline
{\scriptsize (Fig.~\ref{fig:augmentation_variation})} &
All &
$+3.23$ avg. &
$1\times$ &
$1\times$ \\

\midrule

Four epochs \newline
vs. one epoch
{\scriptsize (Fig.~\ref{fig:epochs_variation})} &
All &
$+3.61$ avg. &
$4\times$ &
$1\times$ \\

\midrule

\multirow{4}{=}{Mixed inference \newline
vs. resize-only
{\scriptsize (Fig.~\ref{fig:crop_resize_variation})}} &
DINOv2 &
$+0.19$ &
\multirow{4}{*}{$1\times$} &
\multirow{4}{*}{$6\times$} \\
&
ViT-L CLIP &
$-0.27$ &
&
\\
&
ResNet-50 CLIP &
$-2.90$ &
&
\\
&
EfficientNet-B0 &
$-2.09$ &
&
\\

\midrule

\multirow{4}{=}{Auxiliary multiclass head \newline
vs. binary
{\scriptsize (Fig.~\ref{fig:multihead_variation})}} &
DINOv2 &
$-0.54$ &
\multirow{4}{*}{$\approx 1\times$} &
\multirow{4}{*}{$\approx 1\times$} \\
&
ViT-L CLIP &
$+0.48$ &
&
\\
&
ResNet-50 CLIP &
$-2.76$ &
&
\\
&
EfficientNet-B0 &
$-0.10$ &
&
\\

\midrule

Harmonic replay \newline
vs. full batch retraining
{\scriptsize (Fig.~\ref{fig:cl_next})} &
All &
$-0.91$ avg. &
$0.43\times$ &
$1\times$ \\

\bottomrule
\end{tabular}
\caption{Accuracy--cost summary for the main recommendations. AUROC differences are expressed in percentage points (pp) and are computed from the three-run means reported in the previous tables. Mixed inference uses one resized input plus five crops and is applied to resize-trained models. For replay, the \emph{Next Period} average is reported. The corresponding 95\% bootstrap confidence intervals are reported in Table~\ref{tab:app:cost}.}
\label{tab:uncertainty_cost}
\end{table}

\section{Conclusions}
\label{sec:conclusions}
In this paper, we presented a comprehensive evaluation of the design factors that influence the performance and generalization of deepfake and AI-generated image detectors. Our analysis covered a wide range of training and inference choices, including augmentation strategies, preprocessing pipelines, training duration, multiclass supervision, and continual learning mechanisms.

Our findings separate robust cross-backbone recommendations from architecture-dependent observations. First, aligning the training distribution with realistic degradation is beneficial: using the AI-GenBench \emph{Evaluation} pipeline, which applies multiple JPEG compression passes, consistently outperforms more aggressive augmentation strategies. Second, training for four epochs while maintaining the standard augmentation multiplier ($am=4$) offers an excellent performance–efficiency trade-off. Third, full-image resizing emerges as the most stable and reliable input processing strategy: crop-based or hybrid alternatives never yield a gain beyond the variability across runs, and hybrid inference clearly degrades the performance of the smaller backbones. Finally, direct optimization of the binary objective remains the most robust approach, although a dual-head configuration with an auxiliary multiclass loss can achieve comparable performance in larger models while enabling model attribution.
Beyond the standard batch setting, we examined how detectors can be periodically updated as new generative models become available. Our experiments demonstrate that the proposed \emph{Harmonic replay} strategy achieves performance close to full retraining (on average $0.91$ percentage points lower on the \emph{Next Period}) while significantly reducing computational costs and mitigating the influence of obsolete generators, making it a practical and scalable solution for maintaining detectors up to date with novel generative models. 
By integrating the identified ``best of'' practices on the DINOv2 backbone, we obtained, in a single training run, state-of-the-art performance on the AI-GenBench benchmark. Overall, realistic corruption-aligned augmentation and sufficient training exposure are broadly useful, whereas input fusion and generator-aware supervision should be selected with backbone-specific evidence and cost in mind.

This study has two main limitations. First, all experiments are conducted within AI-GenBench and its fixed evaluation pipeline. AI-GenBench is appropriate to evaluate controlled temporal generalization, but it does not fully cover stronger distribution shifts or deployment pipelines with different corruptions. Secondly, the benchmark focuses on whole-image synthetic content and does not directly consider localized editing and inpainting. Actually, some preliminary experiments have been conducted in \cite{pandolfini2026detecting} to assess the generalization capability of the \emph{best of} recipe to other benchmarks focused on inpainting. Results show that the proposed \emph{best of}, even if not specifically trained on inpainting datasets, performs fairly well if the inpainted part of images is not too small. Future work will extend this systematic analysis to more challenging settings, such as inpainting and localized manipulations.

\appendix
\section{Bootstrap Confidence Intervals}
\label{app:bootstrap}
The tables in this appendix report the nonparametric bootstrap confidence intervals computed on the stored per-item prediction scores of a single run, using $10000$ bootstrap replicates and taking the 2.5th and 97.5th percentiles as $95\%$ bounds. They quantify the uncertainty due to finite test-set sampling and do not capture the variability across random seeds, which is instead reported as mean $\pm$ standard deviation in the main text. For this reason, the point estimates quoted here (single run) can differ slightly from the three-run means reported in the main tables.

\begin{table}[H]
\centering
\footnotesize
\setlength{\tabcolsep}{2.5pt}
\begin{tabular}{lcccc}
\toprule
\textbf{Pipeline} & \textbf{DINOv2} & \textbf{ViT-L CLIP} & \textbf{ResNet-50 CLIP} & \textbf{EfficientNet-B0} \\
\midrule
Baseline & 94.22 [94.04, 94.40] & 90.79 [90.55, 91.03] & 85.21 [84.93, 85.51] & 88.79 [88.54, 89.03] \\
Evaluation & 95.68 [95.53, 95.83] & 94.55 [94.37, 94.74] & 93.22 [93.03, 93.41] & 90.19 [89.96, 90.41] \\
Mild & 95.43 [95.27, 95.59] & 93.53 [93.33, 93.72] & 90.22 [90.00, 90.45] & 89.69 [89.45, 89.92] \\
\bottomrule
\end{tabular}
\caption{Training-time augmentation pipelines. Entries are the single-run point estimate followed by the 95\% bootstrap confidence interval, for the Next Period AUROC averaged over all temporal evaluation steps (\%).}
\label{tab:app:augmentation}
\end{table}

\begin{table}[H]
\centering
\footnotesize
\setlength{\tabcolsep}{2.5pt}
\begin{tabular}{lcccc}
\toprule
\textbf{$am$} & \textbf{DINOv2} & \textbf{ViT-L CLIP} & \textbf{ResNet-50 CLIP} & \textbf{EfficientNet-B0} \\
\midrule
1 & 87.46 [87.20, 87.73] & 79.93 [79.61, 80.26] & 75.92 [75.56, 76.29] & 75.98 [75.62, 76.33] \\
2 & 91.73 [91.52, 91.95] & 86.44 [86.16, 86.72] & 81.77 [81.45, 82.09] & 86.51 [86.24, 86.78] \\
3 & 93.86 [93.68, 94.04] & 89.71 [89.45, 89.96] & 82.65 [82.33, 82.96] & 87.91 [87.65, 88.17] \\
4 & 94.22 [94.04, 94.40] & 90.79 [90.55, 91.02] & 85.21 [84.92, 85.51] & 88.79 [88.54, 89.03] \\
5 & 95.14 [94.99, 95.31] & 91.83 [91.60, 92.06] & 86.54 [86.26, 86.82] & 89.40 [89.16, 89.63] \\
6 & 95.73 [95.57, 95.88] & 92.30 [92.07, 92.52] & 87.46 [87.19, 87.73] & 89.87 [89.64, 90.11] \\
7 & 95.99 [95.84, 96.14] & 92.94 [92.72, 93.14] & 87.10 [86.82, 87.36] & 90.28 [90.05, 90.51] \\
8 & 96.36 [96.22, 96.50] & 93.17 [92.96, 93.38] & 89.41 [89.16, 89.66] & 90.52 [90.29, 90.75] \\
9 & 96.54 [96.40, 96.67] & 93.43 [93.22, 93.64] & 89.12 [88.86, 89.37] & 90.77 [90.54, 90.99] \\
10 & 96.67 [96.53, 96.80] & 93.54 [93.34, 93.75] & 90.17 [89.93, 90.41] & 90.98 [90.76, 91.20] \\
11 & 96.66 [96.52, 96.79] & 93.59 [93.38, 93.79] & 90.02 [89.77, 90.25] & 91.17 [90.96, 91.39] \\
12 & 96.93 [96.80, 97.06] & 93.69 [93.49, 93.90] & 90.56 [90.32, 90.79] & 91.36 [91.14, 91.57] \\
13 & 97.06 [96.93, 97.18] & 93.57 [93.36, 93.78] & 91.16 [90.93, 91.38] & 91.42 [91.20, 91.63] \\
14 & 96.98 [96.86, 97.11] & 93.57 [93.36, 93.77] & 90.93 [90.69, 91.16] & 91.66 [91.45, 91.88] \\
15 & 97.08 [96.95, 97.20] & 93.61 [93.40, 93.81] & 91.77 [91.54, 91.99] & 91.73 [91.52, 91.94] \\
16 & 97.11 [96.98, 97.23] & 93.46 [93.25, 93.67] & 91.44 [91.21, 91.67] & 91.85 [91.64, 92.06] \\
\bottomrule
\end{tabular}
\caption{Augmentation multiplier ($am$), with the number of training epochs fixed to one. Entries are the single-run point estimate followed by the 95\% bootstrap confidence interval, for the Next Period AUROC averaged over all temporal evaluation steps (\%).}
\label{tab:app:am}
\end{table}

\begin{table}[H]
\centering
\footnotesize
\setlength{\tabcolsep}{2.5pt}
\begin{tabular}{lcccc}
\toprule
\textbf{Epochs} & \textbf{DINOv2} & \textbf{ViT-L CLIP} & \textbf{ResNet-50 CLIP} & \textbf{EfficientNet-B0} \\
\midrule
1 & 94.22 [94.04, 94.40] & 90.79 [90.55, 91.03] & 85.21 [84.93, 85.51] & 88.79 [88.54, 89.03] \\
2 & 96.38 [96.25, 96.52] & 93.41 [93.20, 93.61] & 89.20 [88.95, 89.44] & 90.39 [90.16, 90.61] \\
3 & 96.94 [96.81, 97.07] & 93.46 [93.26, 93.67] & 90.68 [90.45, 90.91] & 91.21 [91.00, 91.43] \\
4 & 97.09 [96.97, 97.21] & 93.31 [93.10, 93.53] & 92.04 [91.82, 92.25] & 91.73 [91.52, 91.93] \\
\bottomrule
\end{tabular}
\caption{Training duration (epochs), with $am=4$. Entries are the single-run point estimate followed by the 95\% bootstrap confidence interval, for the Next Period AUROC averaged over all temporal evaluation steps (\%).}
\label{tab:app:epochs}
\end{table}

\begin{table}[H]
\centering
\footnotesize
\setlength{\tabcolsep}{2.5pt}
\begin{tabular}{lcccc}
\toprule
\textbf{Train. / Eval. input} & \textbf{DINOv2} & \textbf{ViT-L CLIP} & \textbf{ResNet-50 CLIP} & \textbf{EfficientNet-B0} \\
\midrule
Resize / Resize & 94.22 [94.04, 94.40] & 90.79 [90.55, 91.03] & 85.21 [84.93, 85.51] & 88.79 [88.54, 89.03] \\
Resize / Mixed & 94.63 [94.45, 94.80] & 90.65 [90.41, 90.90] & 82.81 [82.49, 83.13] & 86.79 [86.52, 87.06] \\
Crop / Mixed & 94.95 [94.78, 95.11] & 87.48 [87.21, 87.76] & 82.72 [82.41, 83.02] & 86.54 [86.27, 86.81] \\
\bottomrule
\end{tabular}
\caption{Training and evaluation input strategies; rows follow the order of Table~\ref{tab:crop_resize_variation} (resize/resize, resize/mixed, crop/mixed). Entries are the single-run point estimate followed by the 95\% bootstrap confidence interval, for the Next Period AUROC averaged over all temporal evaluation steps (\%).}
\label{tab:app:crop_resize}
\end{table}

\begin{table}[H]
\centering
\footnotesize
\setlength{\tabcolsep}{2.5pt}
\begin{tabular}{lcccc}
\toprule
\textbf{Fusion strategy} & \textbf{DINOv2} & \textbf{ViT-L CLIP} & \textbf{ResNet-50 CLIP} & \textbf{EfficientNet-B0} \\
\midrule
Baseline & 94.22 [94.04, 94.40] & 90.79 [90.55, 91.03] & 85.21 [84.93, 85.51] & 88.79 [88.54, 89.03] \\
Sum fusion & 92.87 [92.67, 93.07] & 84.55 [84.27, 84.82] & 82.81 [82.50, 83.12] & 87.86 [87.59, 88.11] \\
Max fusion & 92.31 [92.10, 92.52] & 84.55 [84.26, 84.83] & 82.64 [82.33, 82.95] & 88.67 [88.41, 88.91] \\
\bottomrule
\end{tabular}
\caption{Plain multiclass training with sum and max fusion; rows follow the order of Table~\ref{tab:multiclass_plain}. Entries are the single-run point estimate followed by the 95\% bootstrap confidence interval, for the Next Period AUROC averaged over all temporal evaluation steps (\%).}
\label{tab:app:multiclass}
\end{table}

\begin{table}[H]
\centering
\footnotesize
\setlength{\tabcolsep}{2.5pt}
\begin{tabular}{lcccc}
\toprule
\textbf{Configuration} & \textbf{DINOv2} & \textbf{ViT-L CLIP} & \textbf{ResNet-50 CLIP} & \textbf{EfficientNet-B0} \\
\midrule
Baseline & 94.22 [94.04, 94.40] & 90.79 [90.55, 91.03] & 85.21 [84.93, 85.51] & 88.79 [88.54, 89.03] \\
Separate heads & 93.15 [92.96, 93.35] & 89.03 [88.79, 89.27] & 82.99 [82.68, 83.30] & 88.05 [87.79, 88.30] \\
Stacked heads & 91.17 [90.95, 91.39] & 84.95 [84.69, 85.20] & 77.70 [77.40, 78.01] & 86.08 [85.81, 86.35] \\
Separate heads (aux) & 94.21 [94.03, 94.39] & 92.35 [92.14, 92.55] & 84.09 [83.79, 84.39] & 88.55 [88.30, 88.80] \\
Stacked heads (aux) & 93.66 [93.47, 93.85] & 86.74 [86.49, 87.00] & 81.65 [81.34, 81.96] & 85.65 [85.37, 85.92] \\
\bottomrule
\end{tabular}
\caption{Dual-head training; rows follow the order of Table~\ref{tab:multihead}. Entries are the single-run point estimate followed by the 95\% bootstrap confidence interval, for the Next Period AUROC averaged over all temporal evaluation steps (\%).}
\label{tab:app:multihead}
\end{table}

\begin{table}[H]
\centering
\footnotesize
\setlength{\tabcolsep}{2.5pt}
\begin{tabular}{llcccc}
\toprule
\textbf{Centroids} & \textbf{Fusion} & \textbf{DINOv2} & \textbf{ViT-L CLIP} & \textbf{ResNet-50 CLIP} & \textbf{EfficientNet-B0} \\
\midrule
Baseline & -- & 94.22 [94.04, 94.40] & 90.79 [90.55, 91.03] & 85.21 [84.93, 85.51] & 88.79 [88.54, 89.03] \\
1 & Sum & 86.69 [86.41, 86.96] & 84.14 [83.81, 84.46] & 71.84 [71.44, 72.23] & 79.43 [79.09, 79.76] \\
2 & Sum & 86.35 [86.07, 86.63] & 85.23 [84.92, 85.54] & 70.88 [70.48, 71.27] & 78.63 [78.28, 78.97] \\
3 & Sum & 86.73 [86.45, 87.00] & 84.93 [84.62, 85.24] & 70.48 [70.08, 70.86] & 77.58 [77.23, 77.93] \\
1 & Max & 83.15 [82.86, 83.43] & 83.98 [83.65, 84.31] & 77.71 [77.33, 78.07] & 76.10 [75.75, 76.45] \\
2 & Max & 80.88 [80.58, 81.18] & 84.63 [84.31, 84.95] & 72.10 [71.70, 72.49] & 76.17 [75.83, 76.52] \\
3 & Max & 79.98 [79.67, 80.29] & 83.50 [83.18, 83.82] & 68.06 [67.65, 68.48] & 74.59 [74.23, 74.96] \\
\bottomrule
\end{tabular}
\caption{Distance-based scoring; the first three data rows use sum fusion and the last three max fusion, as in Table~\ref{tab:centroid_variation}. Entries are the single-run point estimate followed by the 95\% bootstrap confidence interval, for the Next Period AUROC averaged over all temporal evaluation steps (\%).}
\label{tab:app:centroid}
\end{table}

\begin{table}[H]
\centering
\footnotesize
\setlength{\tabcolsep}{2.5pt}
\begin{tabular}{lcccc}
\toprule
\textbf{Configuration} & \textbf{DINOv2} & \textbf{ViT-L CLIP} & \textbf{ResNet-50 CLIP} & \textbf{EfficientNet-B0} \\
\midrule
Baseline (reference) & 94.22 [94.04, 94.40] & 90.79 [90.55, 91.03] & 85.21 [84.93, 85.51] & 88.79 [88.54, 89.03] \\
Reset weights & 91.01 [90.79, 91.22] & 86.69 [86.43, 86.96] & 81.08 [80.75, 81.41] & 87.09 [86.83, 87.36] \\
Naive & 91.83 [91.62, 92.03] & 88.36 [88.10, 88.62] & 80.15 [79.81, 80.48] & 87.17 [86.91, 87.43] \\
Replay (CB, 10k) & 92.14 [91.94, 92.34] & 90.66 [90.43, 90.90] & 82.19 [81.88, 82.51] & 87.31 [87.05, 87.57] \\
Replay (CB, 20k) & 93.29 [93.10, 93.48] & 90.96 [90.72, 91.19] & 81.66 [81.33, 81.99] & 87.23 [86.98, 87.49] \\
Replay (harmonic) & 93.76 [93.57, 93.94] & 91.87 [91.65, 92.09] & 84.13 [83.82, 84.43] & 87.03 [86.77, 87.30] \\
\bottomrule
\end{tabular}
\caption{Incremental update strategies, Next Period metric; rows follow the order of Table~\ref{tab:cl_next}. Entries are the single-run point estimate followed by the 95\% bootstrap confidence interval, for the Next Period AUROC averaged over all temporal evaluation steps (\%).}
\label{tab:app:cl_next}
\end{table}

\begin{table}[H]
\centering
\footnotesize
\setlength{\tabcolsep}{2.5pt}
\begin{tabular}{lcccc}
\toprule
\textbf{Configuration} & \textbf{DINOv2} & \textbf{ViT-L CLIP} & \textbf{ResNet-50 CLIP} & \textbf{EfficientNet-B0} \\
\midrule
Baseline (reference) & 99.24 [99.20, 99.28] & 97.81 [97.74, 97.89] & 92.41 [92.27, 92.55] & 95.42 [95.33, 95.51] \\
Reset weights & 97.98 [97.93, 98.03] & 95.79 [95.70, 95.88] & 86.65 [86.49, 86.81] & 93.41 [93.31, 93.51] \\
Naive & 96.50 [96.45, 96.56] & 93.62 [93.53, 93.72] & 82.13 [81.95, 82.30] & 91.92 [91.82, 92.03] \\
Replay (CB, 10k) & 98.20 [98.16, 98.25] & 96.74 [96.65, 96.82] & 88.03 [87.88, 88.19] & 93.12 [93.02, 93.22] \\
Replay (CB, 20k) & 98.83 [98.79, 98.86] & 96.95 [96.87, 97.03] & 87.40 [87.24, 87.56] & 93.43 [93.33, 93.53] \\
Replay (harmonic) & 98.88 [98.85, 98.92] & 97.46 [97.38, 97.54] & 90.36 [90.21, 90.51] & 93.20 [93.10, 93.31] \\
\bottomrule
\end{tabular}
\caption{Incremental update strategies, Past Period metric; rows follow the order of Table~\ref{tab:cl_past}. Entries are the single-run point estimate followed by the 95\% bootstrap confidence interval, for the Past Period AUROC averaged over all temporal evaluation steps (\%).}
\label{tab:app:cl_past}
\end{table}

\begin{table}[H]
\centering
\footnotesize
\renewcommand{\arraystretch}{1.0}
\setlength{\tabcolsep}{3pt}
\begin{tabular}{m{3.6cm} l c}
\toprule
\textbf{Recommendation} & \textbf{Backbone} & \textbf{AUROC $\Delta$ (pp) [95\% CI]} \\
\midrule
Evaluation augmentation vs. baseline & All & $+3.66$ avg. $[+3.23,+4.09]$ \\
\midrule
Four epochs vs. one epoch & All & $+3.79$ avg. $[+3.36,+4.22]$ \\
\midrule
\multirow{4}{=}{Mixed inference vs. resize-only} & DINOv2 & $+0.41$ $[+0.06,+0.76]$ \\
& ViT-L CLIP & $-0.13$ $[-0.62,+0.35]$ \\
& ResNet-50 CLIP & $-2.40$ $[-3.02,-1.80]$ \\
& EfficientNet-B0 & $-2.00$ $[-2.52,-1.48]$ \\
\midrule
\multirow{4}{=}{Auxiliary multiclass head vs. binary} & DINOv2 & $-0.01$ $[-0.37,+0.35]$ \\
& ViT-L CLIP & $+1.56$ $[+1.12,+2.00]$ \\
& ResNet-50 CLIP & $-1.13$ $[-1.72,-0.54]$ \\
& EfficientNet-B0 & $-0.24$ $[-0.73,+0.26]$ \\
\midrule
Harmonic replay vs. full batch retraining & All & $-0.56$ $[-1.04,-0.07]$ \\
\bottomrule
\end{tabular}
\caption{95\% bootstrap confidence intervals of the AUROC differences summarized in Table~\ref{tab:uncertainty_cost}. Differences and intervals are computed from the single-run point estimates of the compared configurations and are reported here for completeness; Table~\ref{tab:uncertainty_cost} reports the same differences computed from the three-run means.}
\label{tab:app:cost}
\end{table}

\vspace{6pt}

\section*{Author Contributions}

\textbf{Lorenzo Pellegrini}: Conceptualization, Methodology, Software, Validation, Investigation, Data Curation, Writing - Original Draft, Writing - Review \& Editing; \textbf{Serafino Pandolfini}: Software, Validation, Data Curation, Writing - Review \& Editing; \textbf{Davide Maltoni}: Conceptualization, Investigation, Resources, Writing - Original Draft, Writing - Review \& Editing, Supervision, Project administration, Funding acquisition; \textbf{Matteo Ferrara}: Conceptualization, Investigation, Resources, Writing - Review \& Editing, Supervision; \textbf{Marco Prati}: Validation, Writing - Review \& Editing; \textbf{Marco Ramilli}: Writing - Review \& Editing, Supervision, Funding acquisition.

\section*{Funding}

We acknowledge, for the first author, the support of the European funds from the Emilia-Romagna Region under the Fse+ 2021-2027 program.

\section*{Data Availability Statement}

Data and code will be available upon acceptance at the following link: \url{https://github.com/MI-BioLab/AI-GenBench}.

\noindent\textbf{Generative AI Disclosure:} No generative AI tools were used in the manuscript preparation process.

\section*{Acknowledgment}

We acknowledge ISCRA for awarding this project access to the LEONARDO supercomputer, owned by the EuroHPC Joint Undertaking, hosted by CINECA (Italy).

\section*{Conflicts of Interest}

The authors declare that they have no known competing financial interests or personal relationships that could have appeared to influence the work reported in this paper.

\printbibliography

\end{document}